\documentclass[manuscript,screen]{acmart}

\usepackage{hyperref}
\usepackage{algorithm}
\usepackage{algpseudocode}
\usepackage{subcaption}
\usepackage{tikz,lipsum,lmodern}
\usepackage[most]{tcolorbox}
\usepackage{multirow}
\usepackage{soul}
\usepackage{xcolor}
\usepackage[framemethod=tikz]{mdframed}
\usepackage{enumitem}

\AtBeginDocument{%
  }

\begin{document}
\setcopyright{none}
\settopmatter{printacmref=true}
\newcommand{\qkm}{\textit{Q4\_K\_M}}
\newcommand{\qo}{\textit{Q4\_0}}

\title[Feasibility and Trade-offs of Language Model Inference at the Edge]{Sometimes Painful but Promising: Feasibility and Trade-offs of On-Device Language Model Inference}

\author{Maximilian Abstreiter}

\email{maximilian.abstreiter@helsinki.fi}
\orcid{0009-0002-9535-2393}

\affiliation{%
  \institution{University of Helsinki}
  \department{Department of Computer Science}
  \city{Helsinki}
  \country{Finland}
}

\author{Sasu Tarkoma}

\email{sasu.tarkoma@helsinki.fi}
\orcid{0000-0003-4220-3650}

\affiliation{%
  \institution{University of Helsinki}
  \department{Department of Computer Science}
  \city{Helsinki}
  \country{Finland}
}

\author{Roberto Morabito}

\email{roberto.morabito@eurecom.fr}
\orcid{0000-0002-4240-9934}

\affiliation{%
  \institution{EURECOM}
  \department{Communication Systems Department}
  \city{Sophia Antipolis}
  \country{France}
}
\affiliation{%
  \institution{University of Helsinki}
  \department{Department of Computer Science}
  \city{Helsinki}
  \country{Finland}
}

\renewcommand{\shortauthors}{Abstreiter et al.}

\begin{abstract}
The rapid rise of Language Models (LMs) has expanded the capabilities of natural language processing, powering applications from text generation to complex decision-making. While state-of-the-art LMs often boast hundreds of billions of parameters and are primarily deployed in data centers, recent trends show a growing focus on compact models—typically under 10 billion parameters—enabled by techniques such as quantization and other model compression techniques. This shift paves the way for LMs on edge devices, offering potential benefits such as enhanced privacy, reduced latency, and improved data sovereignty. However, the inherent complexity of even these smaller models, combined with the limited computing resources of edge hardware, raises critical questions about the practical trade-offs in executing LM inference outside the cloud. To address these challenges, we present a comprehensive evaluation of generative LM inference on representative CPU-based and GPU-accelerated edge devices. Our study measures key performance indicators—including memory usage, inference speed, and energy consumption—across various device configurations. Additionally, we examine throughput-energy trade-offs, cost considerations, and usability, alongside an assessment of qualitative model performance. While quantization helps mitigate memory overhead, it does not fully eliminate resource bottlenecks, especially for larger models. Our findings quantify the memory and energy constraints that must be considered for practical real-world deployments, offering concrete insights into the trade-offs between model size, inference performance, and efficiency. The exploration of LMs at the edge is still in its early stages. We hope this study provides a foundation for future research, guiding the refinement of models, the enhancement of inference efficiency, and the advancement of edge-centric AI systems.
\end{abstract}


\maketitle

\section{Introduction}
\label{sec:intro}

Language models (LMs) have gained significant attention in recent years, particularly following the launch of the ChatGPT service in autumn 2022 \cite{chatgpt}.

Beyond their role in conversational AI and virtual assistants, these models are driving substantial transformations across multiple industries, including healthcare, education, entertainment, and software development \cite{pengHealthcare2023, zhangSimulatingClassroomEducation2024, sweetserGaming2024, belznerSoftware2024}. Leading-edge models like GPT-4o have established new benchmarks for AI performance, excelling in general knowledge evaluation (MMLU), mathematical reasoning (GSM8K), and code generation (HumanEval), while also supporting multimodal interactions \cite{hurst2024gpt}. Even fields such as IoT and edge computing are already integrating or preparing to adopt these advanced technologies.
These advancements come, however, with significant computational and storage demands, which are primarily managed through cloud-based infrastructures. While cloud-hosted inference provides scalability and centralized optimization, it also introduces challenges such as increased network latency, data privacy concerns, and high operational costs.

\begin{figure}[!th] 
\begin{center}
\begin{subfigure}{0.342\textwidth}
        \centering
        \includegraphics[width=\linewidth]{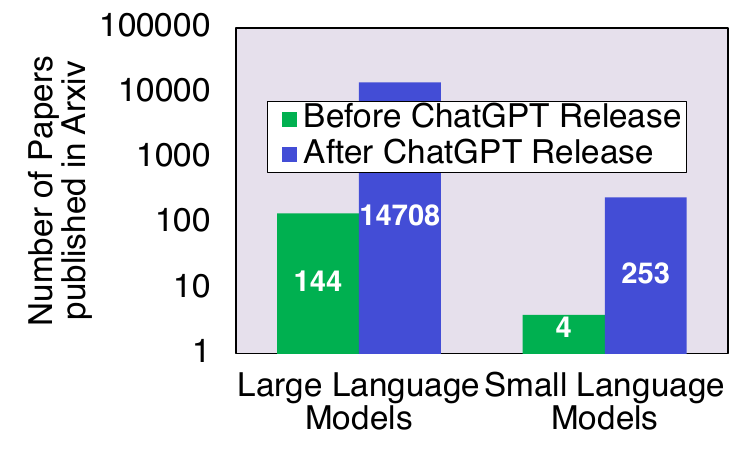}
        \caption{}
        \label{fig:slm_chatgpt}
    \end{subfigure}
    \begin{subfigure}{0.522\textwidth}
        \centering
        \includegraphics[width=\linewidth]{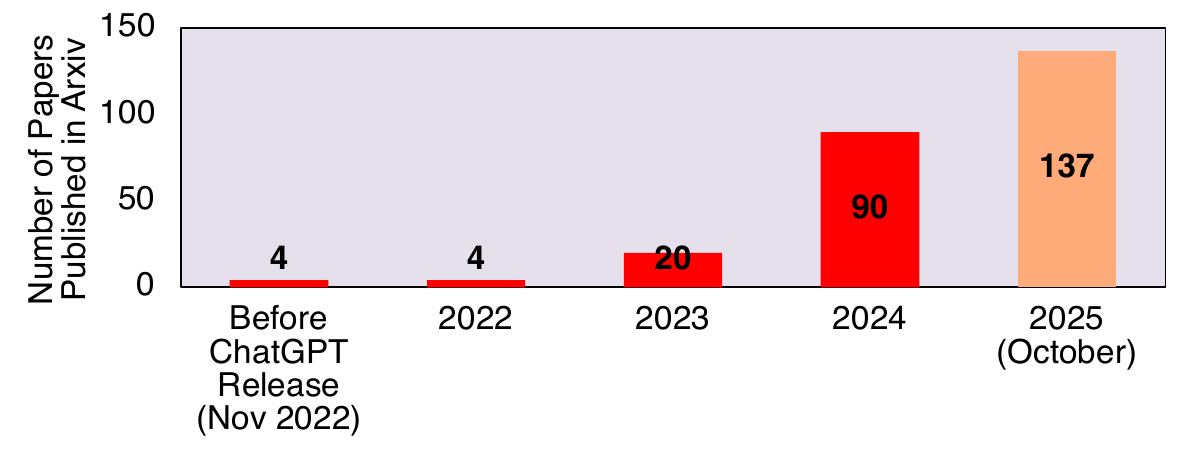}
        \caption{}
        \label{fig:slm_papers}
    \end{subfigure} \hfill
\caption{Comparison of the number of research papers published on arXiv mentioning 'Large Language Model' (LLM) or 'Small Language Model' (SLM) in their title before and after the release of ChatGPT (a) and for SLM over the last years (b), illustrating the rising research interest in edge-optimized AI models.}
\label{fig:paperarxiv}
\end{center}
\end{figure}

To address these issues, there is a growing trend toward deploying LMs at the edge, paralleling earlier shifts in AI inference from centralized data centers to resource-constrained, decentralized environments. Similar to pre-LM AI workloads, efforts to enable LM execution closer to the edge are driven by the need to reduce latency, enhance reliability, and improve user experiences. 
This growing interest in smaller-scale models is also reflected in the research community. As shown in Figure~\ref{fig:paperarxiv}, the number of papers published on arXiv mentioning "Large Language Model" in their titles skyrocketed from 184 (over a decade) to 8,480 in just a few years after ChatGPT's release. While work on Small Language Models (SLMs) remains a more niche area, its trajectory follows a similar trend, with research output rising from just 4 papers before ChatGPT’s release to over 126 in early 2025. This indicates that SLMs are becoming a significant research focus, aligning with the broader push toward efficient, resource-constrained AI models.
It is important to note that there is no strict definition of what differentiates an LLM from an SLM in terms of parameter count. While LLMs often exceed tens or even hundreds of billions of parameters, SLMs typically remain under 10 billion, though this boundary is not universally defined. Nevertheless, the trend toward optimizing models for edge and mobile deployment is clear, reinforcing the need to investigate the feasibility of running LMs on resource-constrained devices.

Despite this increasing interest in edge-centric LMs, however, the question remains: \textbf{\textit{to what extent can resource-constrained devices efficiently execute LM inference?}} 

\begin{figure}[!th] 
\begin{center}
\includegraphics[width=0.8\linewidth]{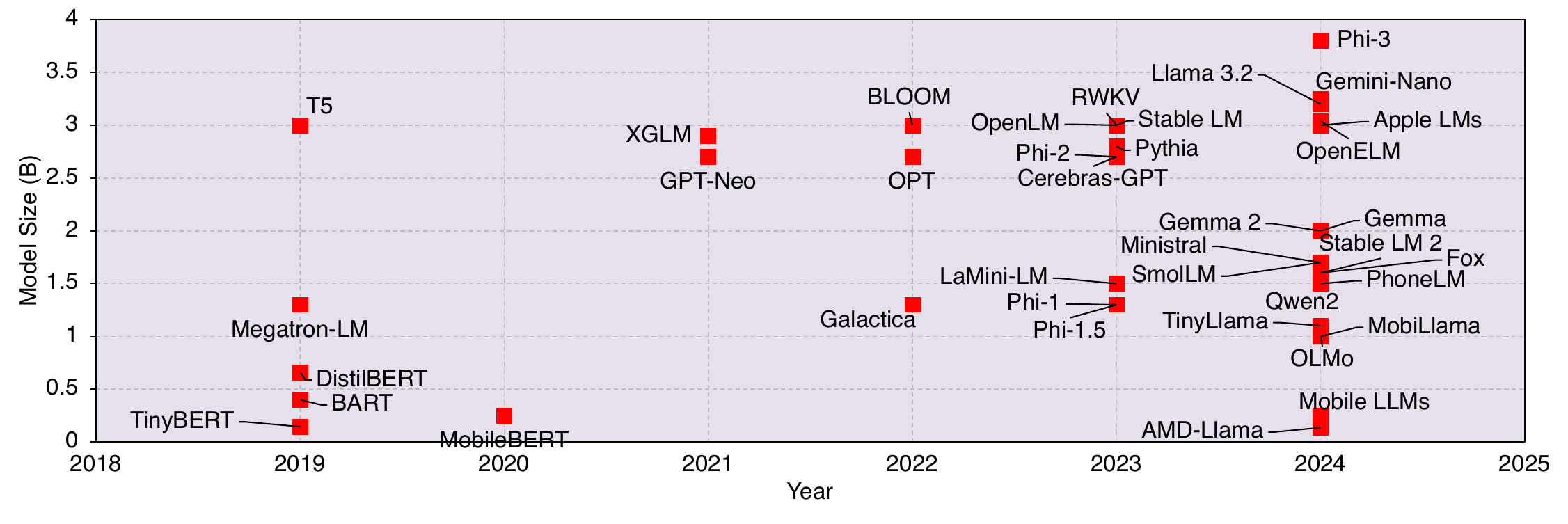}
\caption{Growth of sub-4B parameter language models over recent years, reflecting the increasing trend toward compact models optimized for edge and mobile deployment.}
\label{fig:slm_trend}
\end{center}
\end{figure}

Recent advances in model compression \cite{zhuSurveyModelCompression2024}, such as quantization and knowledge distillation, have enabled the development of more efficient LMs for edge and mobile environments. While quantization reduces memory and computational requirements by lowering parameter precision, distillation transfers knowledge to smaller models with fewer parameters, making them more suitable for resource-constrained settings. Examples include Phi-3.5 \cite{abdinPhi3TechnicalReport2024a}, Llama 3.2 \cite{llama32_annoucement}, and a variety of emerging mobile-optimized LMs. In this respect, the rapid rise of these compact models reflects a broader shift toward efficient, resource-aware AI, as shown in Figure~\ref{fig:slm_trend}, which illustrates the increasing number of sub-4B models releases over recent years–we focus on sub-4B models to maintain clarity, as including larger models would make the visualization overly crowded.

While these advancements improve feasibility for on-device inference, deploying even sub-10B models on edge devices remains computationally challenging. Unlike cloud environments, edge hardware is inherently constrained by limited memory, lower computational power, and energy efficiency concerns. Consequently, the trade-offs between model size, performance, and efficiency must be carefully evaluated to determine how well edge devices can handle LM inference in real-world applications. 

To explore these trade-offs, this paper presents a comprehensive performance evaluation of generative LM inference on two distinct single-board computers (SBCs): the CPU-based Raspberry Pi 5 and the GPU-accelerated NVIDIA Jetson Orin Nano. 
Our study assesses key performance metrics, including memory usage, inference speed, and energy consumption, under different device configurations. Additionally, we investigate the throughput-energy trade-offs, cost implications, usability factors, and qualitative model performance. By quantifying the specific limits imposed by memory and energy constraints, this work wants to offer concrete and practical insights into the feasibility and trade-offs of running LMs at the edge.

Specifically, we explore the following research questions (RQ\#):
\begin{itemize}
    \item[\textbf{(RQ1)}] What are the computational limits of running LMs on edge devices?
    \item[\textbf{(RQ2)}] How do different model configurations (size, quantization) impact performance at the edge?
    \item[\textbf{(RQ3)}] What are the trade-offs between CPU- and GPU-based inference on resource-constrained hardware?
    \item[\textbf{(RQ4)}] How do system configurations (power modes, threading, etc.) affect inference efficiency, including microarchitectural-level performance metrics such as cache misses and context switches?
    \item[\textbf{(RQ5)}] What are the usability constraints, cost implications, and qualitative trade-offs when running LM inference on edge devices?
\end{itemize}

To address these questions, this work aims to provide a structured reference for understanding the performance trade-offs and practical limitations of LM inference on resource-constrained devices. Beyond offering insights for hardware-software co-design in edge AI, our findings contribute to a broader understanding of how edge devices can (or cannot) handle LM workloads effectively. This awareness is relevant for scenarios where distributed edge nodes must collaborate to execute LM inference, helping to inform the design of decentralized AI systems. Additionally, this study provides different insights for AI researchers optimizing inference efficiency, system architects designing multi-node edge deployments, and industry practitioners integrating LMs into latency-sensitive IoT and 5G applications. The key contributions (C\#) of this study are:

\begin{itemize}
    \item[\textbf{(C1)}] Addressing \textbf{(RQ1-3)}, we benchmark 11 generative LMs on two widely used SBCs, analyzing key performance indicators such as memory usage, inference speed, and energy efficiency to quantify the feasibility of edge inference.
    \item[\textbf{(C2)}] Answering \textbf{(RQ2)}, we evaluate the effect of quantization and model scaling on inference efficiency, resource utilization, and model performance, highlighting the trade-offs between accuracy and computational cost.
    \item[\textbf{(C3)}] Directly addressing \textbf{(RQ3)}, we compare CPU-based inference against GPU acceleration, investigating their impact on execution speed, energy consumption, and practical deployment feasibility on edge devices.
    \item[\textbf{(C4)}] Responding to \textbf{(RQ4)}, we analyze the impact of power modes, threading configurations, and system settings, while also evaluating micro-architectural metrics, such as cache misses and context switches, to uncover bottlenecks and efficiency gaps in edge-based LM execution.
    \item[\textbf{(C5)}] Informed by \textbf{(RQ5)}, we assess the practical challenges of running LMs at the edge, including inference cost analysis, qualitative benchmarking of the models, and aspects such as usability related to real-world applicability, providing a broader perspective beyond raw performance metrics.
\end{itemize}

The remainder of this paper is structured as follows: Section~\ref{sec:background} provides background information on enabling technologies such as transformers and inference engines. Section~\ref{sec:related} presents a selection of closely related work. Section~\ref{sec:methods} details the experimental methodology, while Section~\ref{sec:results} presents the empirical results. In Section~\ref{sec:discuss}, these results are further discussed and contextualized, leading to the final conclusions in Section~\ref{sec:conclusion}.

\section{Background}
\label{sec:background}

This section provides an overview of the key concepts relevant to our study. We first describe the architecture of LMs, including the transformer architecture, key-value cache (KV-cache) utilization, and model compression techniques such as quantization. Then, we introduce inference backends commonly used to execute LM inference on edge devices, which play an important role in determining computational feasibility.

\textbf{Transformer Architecture and Large Language Models.} The transformer architecture \cite{vaswani_attention_2017} was a breakthrough discovery, and most state-of-the-art LMs are still based on it. Its success stems from the multi-head attention mechanism, which does not rely on convolutions or recurrences, unlike previously established architectures. The attention mechanism computes key, value, and query vectors, where the key and value vectors of each token can be cached in the KV-cache and reused in subsequent forward passes. A token is the fundamental unit of text that language models process. Tokens can represent entire words, subwords, or even individual characters, depending on the tokenization method used. Tokenization is the process of dividing a given text into discrete units (tokens). The KV-cache stores key and value vectors for all tokens within the current context, where the context refers to the sequence of previous tokens that still influence the generation of the next token. The maximum number of tokens that can be stored in the context is referred to as the context size. Because of the KV-cache, the generation of the first token exhibits different computational characteristics compared to subsequent tokens. As a result, generative transformer inference is typically divided into two phases: the \textit{prefill phase}, also called initialization or prompt processing phase, and the \textit{generation phase}. 

In the prefill phase, the model calculates the key and value vectors for every input token and then stores them in the KV-cache. This process is computationally expensive, making the prefill phase compute-bound. Conversely, in the generation phase, these cached vectors are reused, eliminating the need for recomputation. As a result, this phase is memory-bound, as it involves frequent memory accesses. Without a KV-cache, the time complexity of the attention layer is quadratic with respect to sequence length, whereas utilizing a KV-cache reduces the complexity to linear.

The original transformer architecture follows an encoder-decoder approach \cite{cho_encoderdecoder_2014}, consisting of two independent neural networks interacting with each other. The encoder computes a variable-length sequence and encodes it into a fixed-length vector. The decoder then takes this vector as input and generates a variable-length sequence as output. In the context of language processing, these variable-length sequences correspond to \textit{tokenized} text sequences. Over time, new model variants emerged, using only encoders or only decoders. Encoder-only models tend to perform better for tasks that require language understanding (e.g., sentiment analysis), but they cannot generate text independently. A notable set of encoder-only models belongs to the BERT-family \cite{devlin_bert_2019}. Decoder-only models are typically well-suited for tasks involving text generation. Examples of this model class include GPT-4 \cite{openai_gpt-4_2024} or Llama 3 \cite{grattafioriLlama3Herd2024}.

Due to the sheer size of LMs, various model compression techniques have been proposed to improve efficiency, one of the most widely used being quantization. Quantization reduces the model size by lowering the precision of model parameters \cite{lin_awq_2024, frantar_gptq_2023, xiao_smoothquant_2023}. Some quantization techniques focus solely on quantizing model weights, such as GPTQ \cite{frantar_gptq_2023} and k-quants \cite{k-quants}. Other approaches, such as SmoothQuant \cite{xiao_smoothquant_2023}, also quantize model activations, further reducing computational complexity. Quantization techniques can further be classified into \textit{post-training} quantization and \textit{quantization-aware training}. Post-training quantization (e.g., GPTQ, k-quants, SmoothQuant) applies quantization after a model has been fully trained. Quantization-aware training (e.g., BitNet \cite{wang_bitnet_2023, ma_era_2024}) incorporates quantization during training, reducing quality loss. However, it requires training from scratch, making it computationally expensive and inapplicable to pre-trained models.
Some post-training quantization techniques embed limited re-training rounds to mitigate quality loss. These are referred to as n-shot quantization, where n denotes the number of re-training iterations. For example, GPTQ is a 1-shot quantization technique, while 0-shot quantization techniques (e.g., certain k-quants) require no re-training at all.

\textbf{Inference Backends for LMs.} An inference backend is a software tool required to execute AI model inference on a device. Various backends support LM inference across different hardware architectures, enabling execution on both CPU- and GPU-based systems. One of the most widely used inference backends for LMs is \texttt{llama.cpp} \cite{llama-cpp}, known for its broad hardware support and ability to run models on both CPUs and GPUs. This backend system was chosen for this work, because of its flexibility and compatibility with various LM architectures. Another popular backend is \texttt{MLC LLM} \cite{mlc-llm}, which offers a similar range of hardware and model support as llama.cpp. Additionally, it supports various quantization schemes \cite{k-quants}, making it well-suited for optimized on-device inference. \texttt{picoLLM} \cite{picollm} is another alternative, providing wide hardware support but offering a more limited model selection than MLC LLM and llama.cpp. It supports quantization and claims that its X-bit quantization significantly outperforms GPTQ \cite{frantar_gptq_2023}, which is one of the most widely used post-training quantization techniques. For on-device inference, another backend is \texttt{TinyChat} \cite{tinychat}, which supports Activation-aware Weight Quantization (AWQ) \cite{lin_awq_2024} and SmoothQuant \cite{xiao_smoothquant_2023}. However, it currently supports fewer models than llama.cpp or MLC LLM and does not yet enable inference on mobile devices. Finally, Google's \texttt{MediaPipe LLM Inference API} \cite{mediapipellm} enables LM inference capabilities for mobile phones and web applications. However, it currently supports only a limited selection of models.

\section{Related Work}
\label{sec:related}

Benchmarking the performance of deep neural network (DNN) models on devices is not a new topic, and several benchmark suites have been proposed over time, such as AI Benchmark \cite{ignatov_ai-benchmark_2019}, EmBench \cite{almeida_embench_2019} and MLPerf \cite{reddi_mlperf_2020}. Among these, MLPerf has emerged as an industry standard for quantifying device performance for DNN training and inference across different DNN tasks, such as image classification or text summarization.

\begin{table}[!th]
\footnotesize
\centering
\caption{Comparison with closely related papers}
\label{tab:paper_comparison}
\begin{tabular}{l l c c c c c c }
& & Present work  & \cite{laskaridis_melting_2024} & \cite{nezamiGenerativeAIEdge2024a}  & \cite{liLargeLanguageModels2024a} &  \cite{dharEmpiricalAnalysisResource2024} & \cite{sarkar_bert-on-edge_2024}  \\ 
\midrule
\multirow{3}{*}{Device Types}& SBC (CPU)                         & \checkmark & $\times$ & \checkmark & $\times$ & \checkmark & \checkmark \\
                             & SBC (GPU)                         & \checkmark & \checkmark & $\times$ & $\times$ & $\times$   & \checkmark  \\
                             & Mobile Phones                     & $\times$ & \checkmark & $\times$ & \checkmark & $\times$   & $\times$  \\
\midrule
\multirow{4}{*}{Metrics} & Latency/Throughput                    & \checkmark & \checkmark & \checkmark & \checkmark & \checkmark & \checkmark  \\
                         & Energy                                & \checkmark & \checkmark & $\times$ & $\times$     & $\times$   & \checkmark  \\
                         & Memory                                & \checkmark & \checkmark & \checkmark & \checkmark & \checkmark & \checkmark  \\
                         & Micro-architectural Metrics                     & \checkmark & $\times$   & $\times$ & $\times$     & \checkmark & $\times$\\
\midrule
\multirow{3}{*}{Model Architectures} & encoder-only              & $\times$ & $\times$     & $\times$ & $\times$ & $\times$ & \checkmark \\
                                     & decoder-only              & \checkmark & \checkmark & \checkmark & \checkmark & \checkmark & $\times$ \\
                                     & encoder-decoder           & \checkmark & $\times$ & $\times$ & $\times$ & $\times$ & $\times$ \\
\midrule
\multirow{4}{*}{Discussion Points} & Configurations              & \checkmark & (\checkmark) & $\times$ & $\times$ & $\times$ & $\times$ \\
                                   & Throughput-Energy Trade-off & \checkmark & $\times$ & $\times$ & $\times$ & $\times$ & $\times$ \\
                                   & Usability                   & \checkmark & \checkmark & $\times$ & $\times$ & $\times$ & $\times$ \\
                                   & Cost Analysis               & \checkmark & $\times$ & $\times$ & $\times$ & $\times$ & $\times$ \\

\end{tabular}
\end{table}

Recent studies have investigated the performance of transformer-based LMs on edge devices. A comparative overview of these works is provided in Table~\ref{tab:paper_comparison}. Sarkar et al. \cite{sarkar_bert-on-edge_2024} analyzed the performance of BERT-family models \cite{devlin_bert_2019} on multiple SBCs across various NLP tasks. Their work also examined the impact of pruning on performance and accuracy. In contrast, our study focuses on generative models rather than encoder-only architectures and employs quantization as the primary model compression technique. Other works that focus on generative transformer models include \cite{liLargeLanguageModels2024a}, which analyzes LM inference on mobile phones, differing from our focus on SBCs. \cite{dharEmpiricalAnalysisResource2024}, and \cite{nezamiGenerativeAIEdge2024a}, which evaluates LM inference on SBCs but lack energy consumption measurements and do not discuss usability considerations for LM deployment at the edge. 
The study most closely related to our work is MELTing Point \cite{laskaridis_melting_2024}, which also investigates LM inference feasibility on edge devices, analyzing computational performance and energy usage. However, their study primarily focuses on mobile phones and GPU-accelerated edge devices, whereas our work extends the analysis to CPU-based inference on SBCs. MELTing Point is also the only prior work that discusses usability in detail based on its results.
Furthermore, none of these studies explores how inference and system parameters—such as thread count and CPU governor settings—affect performance metrics. MELTing Point partially addresses this aspect by comparing different power modes on NVIDIA Orin devices, but does not generalize it to broader SBC hardware. Additionally, our study is one of the few to evaluate micro-architectural performance metrics (e.g., cache misses, context switches) in the context of LM inference on edge devices—the only other known work that considers this aspect is Dhar et al. \cite{dharEmpiricalAnalysisResource2024}. Finally, our work stands out by providing an in-depth discussion of energy-throughput trade-offs across different configurations and conducting a monetary cost analysis based on energy consumption during LM inference at the edge—an aspect not explored in previous studies. 

\section{Methodology}
\label{sec:methods}

\begin{table}[!th]
\footnotesize
\centering
\caption{Hardware characteristics of evaluated devices}
\label{tab:devices}
\begin{tabular}{r l l}
& Raspberry Pi 5 & Jetson Orin Nano Developer Kit \\ 
\midrule
CPU & Arm Cortex-A76 (4 cores) & Arm Cortex-A78AE (6 cores) \\
GPU & VideoCore VII & NVIDIA Ampere (1024 CUDA cores) \\
RAM & 8 GB LPDDR4X RAM & 8 GB of LPDDR5 RAM \\
Storage & 128 GB SanDisk High Endurance MicroSD & 128 GB SanDisk High Endurance MicroSD \\
OS & Raspberry Pi OS (Debian-based) & NVIDIA Jetson Linux 36.3 (Ubuntu-based) \\
\end{tabular}
\end{table}

As mentioned in Section~\ref{sec:intro}, for this study, we evaluated two widely used SBCs: a Raspberry Pi 5 (abbr. \textit{RPi 5}) \cite{raspberrypiltdRaspberryPi52025} and an NVIDIA Jetson Orin Nano Developer Kit (abbr. \textit{Orin}) \cite{nvidiacorporationNVIDIAJetsonOrin2024, nvidiacorporationJetsonOrinNanoDevKit2024}. These devices were chosen as they represent widespread and accessible platforms for CPU- and GPU-based edge computing. We deliberately excluded earlier versions of these devices, such as the Raspberry Pi 4 or the original NVIDIA Jetson Nano, as they were either unable to execute many experiments due to out-of-memory failures, or they did not provide enough computational power to execute LM inference at satisfying speeds \cite{dharEmpiricalAnalysisResource2024}, making them impractical for real-world LM inference. Conversely, we did not consider significantly more powerful edge devices, such as the Jetson AGX Orin, because its power supply requirements (19V, 90W) and peak power draw ($\sim$60W) place it closer to server-class hardware rather than the class of energy-constrained, low-power SBCs targeted in this study \cite{agx}. With this choice, we wanted to ensure that our evaluation captures realistic trade-offs between CPU- and GPU-accelerated LM inference on practical edge hardware while remaining within the power and resource limitations typical of IoT and edge deployments. 

A summary of the hardware characteristics of both devices is provided in Table~\ref{tab:devices}.
To prevent thermal throttling, the RPi 5 was equipped with a Joy-it RB-Heatsink5, as overheating was observed during initial experiments. On the RPi 5, inference was exclusively CPU-based, while on the Orin, we evaluated both CPU and GPU inference. 

For each device, multiple configurations were tested to analyze the impact of system settings on LLM inference performance. For CPU-based inference, the number of threads was varied from a single thread up to the maximum number of available CPU cores on each device. Additionally, three different CPU frequency scaling governors were evaluated. The \textit{powersave} governor forces the CPU to always run at its minimum frequency, reducing power draw at the cost of potential performance degradation. The \textit{performance} governor keeps the CPU at its maximum frequency at all times, ensuring the highest possible computational throughput. Finally, the \textit{ondemand} governor dynamically adjusts the CPU frequency based on current workload demands, balancing power efficiency and performance; this was the default setting on both devices. For GPU-based inference, the Orin was tested under two predefined power modes: \textit{7W}, which operates within a lower power envelope and may introduce performance constraints, and \textit{15W} mode, which allows higher power draw for improved GPU performance. Additionally, a third setup was evaluated, where both the CPU and GPU frequencies were set to their maximum limits within \textit{15W} mode. This configuration, referred to as \textit{MAX} mode, is used consistently throughout the remainder of this paper.

A single input prompt was used for evaluation, consisting of a summary task of a short story generated by ChatGPT. Due to differences in tokenization methods across models, the number of input tokens varied between 74 and 92, depending on the model. The number of output tokens was fixed at 100 to ensure consistency across experiments, allowing for controlled comparisons across models and configurations. To maintain comparability and repeatability, the context size was set to 512 tokens, as this was the lowest context size for which all evaluated models had been trained. These design choices were made to standardize the experimental setup while allowing fair performance comparisons across different models. While real-world applications may involve variable-length generations, evaluating inference performance across a fixed output length enables a systematic and reproducible assessment of computational trade-offs. Exploring how these trade-offs evolve at higher output scales remains an interesting direction for future work.

\begin{table}[!th]
\footnotesize
\centering
\caption{Overview of evaluated models}
\begin{tabular}{lrrrr}
\label{tab:models}
Model name & Parameters (B) & Architecture & Abbreviation & Reference \\ 
\midrule
Qwen 2 0.5B    & 0.5 & decoder-only    & Q2-0.5B & \cite{yangQwen2TechnicalReport2024}   \\
Flan T5 Large  & 0.8 & encoder-decoder & FT5-L   & \cite{flant5}                         \\
Llama 3.2 1B   & 1.2 & decoder-only    & L3.2-1B & \cite{llama32_annoucement}            \\
Qwen 2 1.5B    & 1.5 & decoder-only    & Q2-1.5B & \cite{yangQwen2TechnicalReport2024}   \\
Gemma 2 2B     & 2.6 & decoder-only    & G2-2B   & \cite{gemma_2024}                     \\
Llama 3.2 3B   & 3.2 & decoder-only    & L3.2-3B & \cite{llama32_annoucement}            \\
Phi 3.5 mini   & 3.8 & decoder-only    & P3.5-M  & \cite{abdinPhi3TechnicalReport2024a}  \\
Yi 1.5 6B      & 6.1 & decoder-only    & Y1.5-6B & \cite{aiYiOpenFoundation2025}         \\
InterLM 2.5 7B & 7.7 & decoder-only    & I2.5-7B & \cite{cai2024internlm2}               \\
Llama 3.1 8B   & 8.0 & decoder-only    & L3.1-8B & \cite{grattafioriLlama3Herd2024}      \\
Gemma 2 9B     & 9.2 & decoder-only    & G2-9B   & \cite{gemma_2024}                     \\
\end{tabular}
\end{table}

The LMs under evaluation were selected based on their rankings in the Hugging Face Open LLM Leaderboard 2 \cite{hf_llm_leaderboard}. To ensure a representative selection, we chose the highest-ranked model for each 1-billion-parameter range up to 9 billion parameters. Instruction fine-tuned base models were prioritized, even in cases where task-specific fine-tuned models ranked higher on the leaderboard. Models in the 4–6 billion parameter range were excluded from evaluation, as the only available models in this range were significantly lower ranked than several smaller models, making their inclusion less meaningful for comparative analysis. For models in the 0–1 billion parameter range, we evaluated two models due to differences in architecture. The highest-ranked model, Flan T5 Large, follows an encoder-decoder architecture, whereas all other evaluated models use a decoder-only architecture. To enable a fair comparison of architectural differences at a similar model scale, we also included the second-highest-ranked model, Qwen 0.5B. Additionally, we evaluated Llama 3.2 1B and 3B \cite{llama32_annoucement}, which were released during the course of this study. Meta claims that these models have been optimized for Arm processors \cite{llama32_annoucement}, making them particularly relevant for our evaluation. For each model, we tested two quantization schemes, \qo{} and \qkm{} \cite{k-quants}. These schemes were chosen based on prior research, which has demonstrated that quantizing below 4 bits in blocking-based post-training quantization results in a significant drop in model quality \cite{dettmers_4bit_2023, laskaridis_melting_2024}. The models were quantized using the llama-quantize tool from llama.cpp. An overview of all evaluated models is provided in Table~\ref{tab:models}, which also includes abbreviations for model names that will be used in some figures throughout the paper for space efficiency.

All experiments were conducted using llama.cpp \cite{llama-cpp} (version 4501) as the inference engine. The experimental workflow was automated through a Bash script to ensure consistency across runs. For each evaluated model and configuration, five runs were performed to enhance statistical reliability. To maintain repeatability, all memory caches and swap memory were cleared before the first run of each experiment. Additionally, a 30-second pause was introduced between runs to allow the devices to cool down, preventing thermal effects from impacting results. To further ensure consistent thermal conditions, the fans on both devices were set to full speed throughout the duration of the experiments.

To measure various performance metrics, we utilized \textit{perf}, a built-in Linux kernel profiler, along with a custom C script that retrieved system metrics from sysfs files. Current and voltage measurements on the RPi 5 were obtained from the power management integrated circuit, using code adapted from the vcgencmd tool \cite{vcgencmd}. While this approach is widely used for energy measurements in edge computing, existing studies highlight that its accuracy and reliability are lower than those obtained via external hardware-based power measurement solutions \cite{shalavi2023accurate}. However, the same study showed that on-device sensor data can be calibrated with a simple linear regression model to estimate the real power draw accurately. We adopted this methodology and achieved a mean absolute percentage error below 2.5\% for both devices on our test data, spanning various workloads with different power profiles. To obtain ground-truth test and training data we used an external Monsoon High Voltage Power Monitor \cite{MonsoonHVPM}. Our evaluation focused on estimating the mean power draw for time periods longer than 100ms, aligned with the shortest time periods of interest in our study. We note that this method would be unsuitable for measuring short-time power spikes due to low sampling rate and increasing inaccuracy for shorter time periods. However, the present work focuses on energy consumption over longer time periods, allowing us to use this method to simplify the experimental setup without significant drop in precision. 
Measurements were recorded every 10 ms using perf, while the custom script implemented a 10 ms sleep interval between measurements. However, due to differences in execution time, the effective measurement intervals varied. On the Orin, the observed measurement interval was typically 15–20 ms, whereas on the RPi 5, it was 30–35 ms, primarily due to the long execution time of current and voltage measurements. We also tested shorter sleep intervals, but these caused significant interference with llama.cpp, distorting the results. Thus, the chosen interval provided a balance between measurement accuracy and minimal impact on inference performance.

\begin{table}[!t]
\small
\centering
\caption{Overview of qualitative downstream task benchmarks}
\begin{tabular}{lrrr}
\label{tab:benchs}
Benchmark & Type & Used Split & Split Sample Size \\ 
\midrule
HellaSwag & Commonsense Reasoning & validation & 10042\\
Winogrande & Commonsense Reasoning & validation & 1267\\
ARC easy  & Question Answering & test & 2376 \\
ARC challenge  & Question Answering & test & 1171 \\
TruthfulQA  & Question Answering & test & 817 \\
\end{tabular}
\end{table}

To assess model quality, we evaluated five different benchmarks using llama-perplexity, a utility included with llama.cpp for performing various benchmarks. First, perplexity was measured on the WikiText-2 test dataset \cite{wikitext2}. Perplexity quantifies a model’s uncertainty in predicting the next token, where lower values indicate higher confidence and better predictive performance and one is the optimal achievable value. In addition to perplexity, models were evaluated on four downstream-task benchmarks (summarized in Table \ref{tab:benchs}): \emph{(i)} HellaSwag \cite{zellers_hellaswag_2019}, \emph{(ii)} Winogrande \cite{sakaguchiWinoGrandeAdversarialWinograd2020}, \emph{(iii)} AI2 Reasoning Challenge (ARC) \cite{clark_arc_2018} divided into easy and challenge subsets, and \emph{(iv)} TruthfulQA \cite{lin_truthfulqa_2022}. One task from the ARC challenge dataset was removed due to special characters causing failures in llama-perplexity. For HellaSwag and Winogrande, the validation dataset splits were used since only unlabeled test splits were available. The Flan T5 Large model was excluded from the qualitative assessment due to compatibility issues with llama-perplexity. All qualitative benchmarks were conducted in a zero-shot setting (i.e., without model fine-tuning on these datasets) using a context size of 512.

\section{Evaluation Results}
\label{sec:results}

This section presents the experimental results. As described earlier, five runs were conducted for each model and configuration, with only runs two to five reported, as the first run served as a warm-up phase. However, differences between the first run and subsequent runs were minimal, except during the load phase, where extended loading times were observed. This phase refers to llama.cpp initializing and loading the model, and its impact will be discussed in the relevant subsections. To compare how different configurations—such as thread count, CPU governor, and power mode—affect various performance metrics, we present detailed results for a single model. Phi 3.5 Mini was chosen as the representative model since it is closest to the center of the evaluated parameter range and did not exhibit significant outlier behavior in any metric. While most models followed similar trends across configurations, substantial outliers are highlighted separately. Despite models being similarly affected by configuration changes, their absolute performance levels varied. These differences are analyzed by presenting results for all models under a single selected configuration: \emph{(i)} CPU governor set to \textit{ondemand} (default setting), \emph{(ii)} Power mode 15W (default setting), and \emph{(iii)} Thread count as maximum available, except for generation throughput and energy consumption, where the optimal thread count was used per model and device. 

Using the maximum thread count caused significant performance degradation in the generation phase, particularly for \qo{} quantization. To address this, we leveraged llama.cpp’s capability to assign separate thread counts for the prefill and generation phases, optimizing inference efficiency without negatively impacting system performance. Unlike thread count, modifying CPU governors and power modes affects the entire system, so we opted to keep them at their default settings to maintain consistency across experiments.

\subsection{Memory Footprint Analysis}
\label{sub:mem}

\begin{figure}[!th]
    \centering
    \begin{subfigure}{0.3\textwidth}
        \centering
        \includegraphics[width=\linewidth]{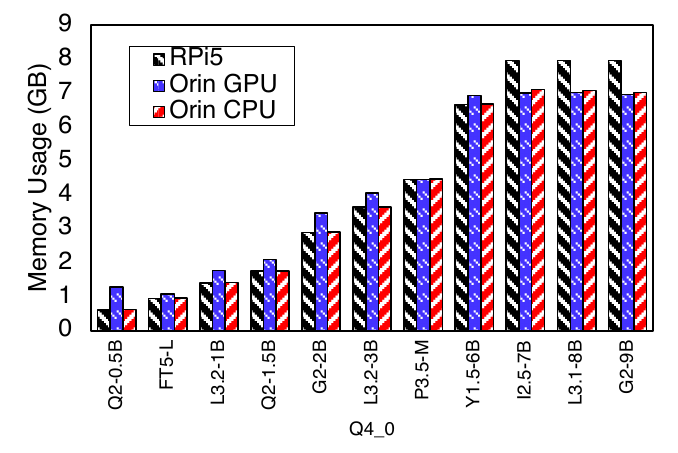}
        \caption{}
        \label{fig:mem_k}
    \end{subfigure}
    \begin{subfigure}{0.3\textwidth}
        \centering
        \includegraphics[width=\linewidth]{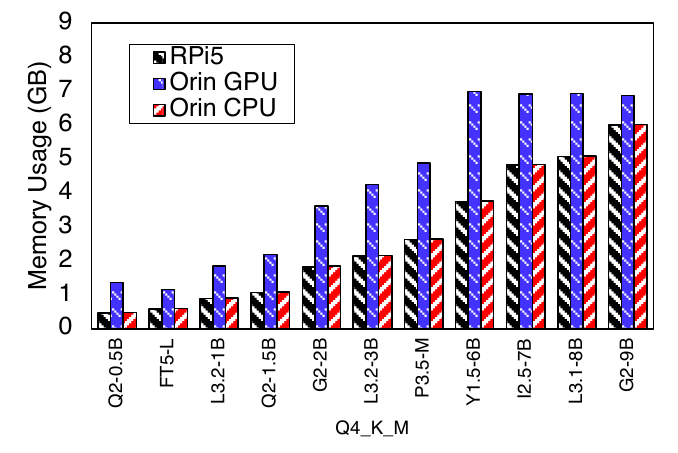}
        \caption{}
        \label{fig:mem_0}
    \end{subfigure} \hfill
    
    \caption{Peak memory usage for all devices and quantization schemes}
    \label{fig:mem_all}
\end{figure}

\begin{figure}[!th]
    \centering
        \begin{subfigure}{\textwidth}
        \centering
        \includegraphics[width=1\linewidth]{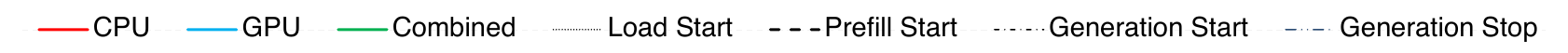}
    \end{subfigure} \hfill
    
    \begin{subfigure}{0.33\textwidth}
        \centering
                \includegraphics[width=\linewidth]{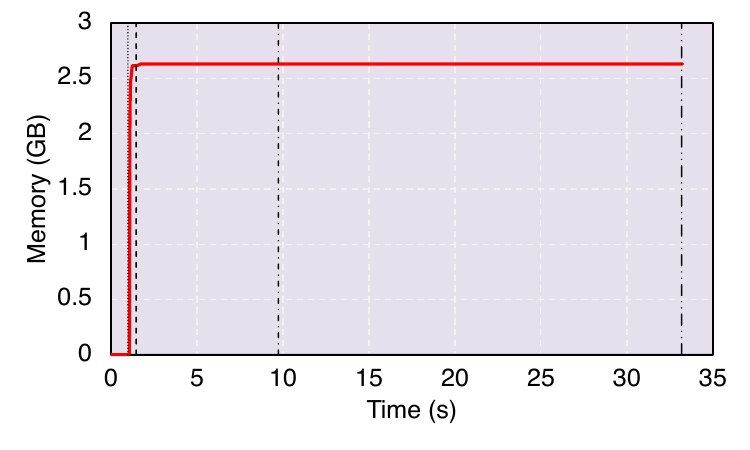}
        \caption{RPi 5, Q4\_K\_M}
        \label{fig:mem_line_pi5_q4km}
    \end{subfigure}
    \begin{subfigure}{0.33\textwidth}
        \centering
        \includegraphics[width=\linewidth]{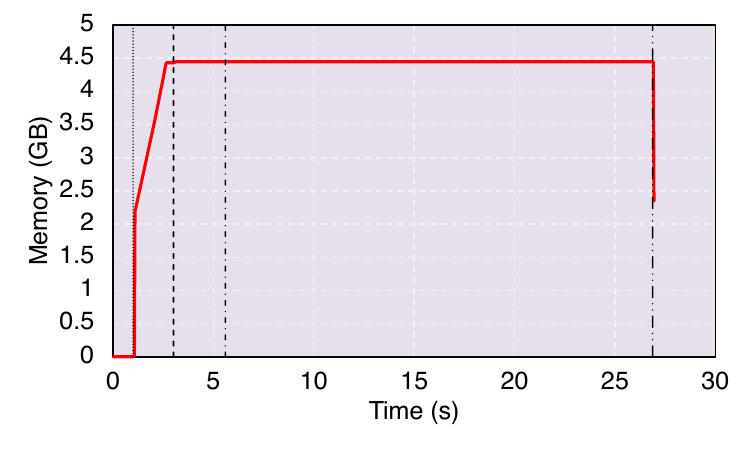}
        \caption{RPi 5, Q4\_0}
        \label{fig:mem_line_pi5_q40}
    \end{subfigure} \hfill
    
    \begin{subfigure}{0.33\textwidth}
        \centering
                \includegraphics[width=\linewidth]{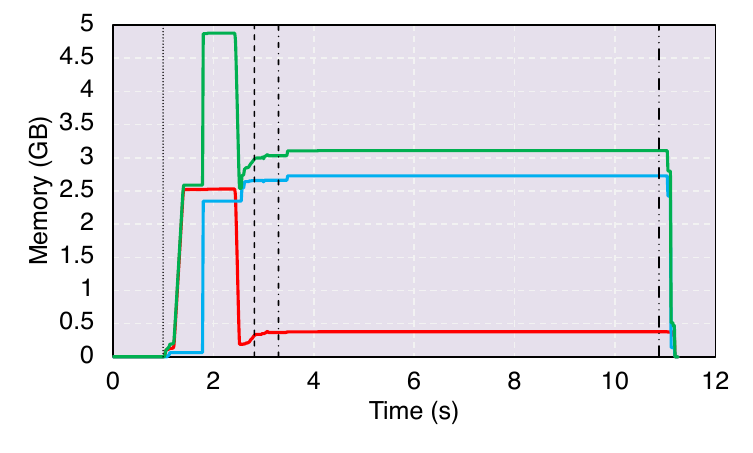}
        \caption{Orin GPU,  Q4\_K\_M}
        \label{fig:mem_line_orin_gpu_q4km}
    \end{subfigure}
    \begin{subfigure}{0.33\textwidth}
        \centering
                \includegraphics[width=\linewidth]{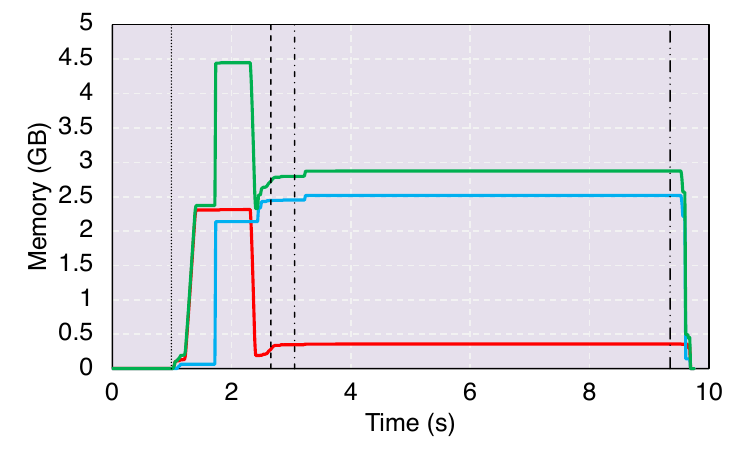}
        \caption{Orin GPU, Q4\_0}
        \label{fig:mem_line_orin_gpu_q40}
    \end{subfigure} \hfill
    \caption{Memory footprint timeline of one example run for RPi 5 (ondemand, 4 threads), Orin GPU (15 W) and both quantization schemes}
    \label{fig:mem_lines_all}
\end{figure}

No significant differences in memory usage were observed across different system configurations. Therefore, this subsection focuses on model- and device-level comparisons.

As shown in Figure~\ref{fig:mem_all}, CPU inference with \qkm{} quantization exhibited an approximately linear relationship between peak memory usage and model size. This behavior is expected, as model weights account for most of the memory footprint. For GPU inference, a larger memory footprint was observed. This is primarily due to the shared main memory between the CPU and GPU on the Orin, leading to additional memory overhead. While the peak GPU memory usage was comparable to CPU inference, the CPU also allocated memory, further increasing overall memory consumption. This effect was particularly pronounced during the load phase, when the model was transferred to the GPU, as seen in Figure~\ref{fig:mem_lines_all}. During this phase, both GPU and CPU memory usage peaked simultaneously before the CPU memory allocation decreased, but still induced additional overhead. Another notable observation in GPU inference was the flattening of memory usage for models larger than Phi 3.5 Mini. This was due to the device exhausting its available memory. Despite having 8 GB of nominal memory, the usable memory was limited to below 7 GB, as the operating system and background processes collectively occupied over 1 GB. 

In contrast, \qo{} quantization exhibited several notable differences in memory usage. First, GPU inference with \qo{} quantization resulted in a slightly lower peak memory footprint for most models, whereas CPU inference showed a significantly higher memory footprint than \qkm{} quantization, as seen in Figure~\ref{fig:mem_all}. The increased memory footprint for CPU inference may seem counterintuitive, as \qo{}-quantized models are smaller than \qkm{}-quantized models. However, this behavior is likely due to the restructuring of model weights during the load phase to leverage optimized matrix multiplication kernels on the Arm CPU \cite{optimize_kernels}. This restructuring process appears to require additional memory, which remains allocated throughout inference, as observed in Figure~\ref{fig:mem_lines_all}. This effect is not present in GPU inference, as model weights are not realigned when loaded onto the GPU. Despite this, for most models, GPU inference still had the highest memory footprint. However, for models larger than Yi 1.5 6B, CPU inference with \qo{} quantization required more memory than GPU inference. Additionally, for Phi 3.5 Mini, memory usage was nearly identical between CPU and GPU inference. A second key observation is that the flattening of memory usage seen in GPU inference with \qkm{} quantization for larger models was also present in CPU inference with \qo{} quantization. As before, this is due to memory exhaustion. The RPi 5's effective memory limit was higher than the Orin's, as its operating system consumed fewer memory resources. However, unlike with \qkm{} quantization, this memory saturation effect did not appear for Yi 1.5 6B under \qo{} quantization. 

It is worth noting that larger context sizes beyond 512 tokens were also tested. As the context size increases, the memory allocated for the KV-cache grows correspondingly. For extensive context sizes, the KV-cache memory footprint can even exceed the memory required for model weights. Even with a moderate context size of 1028, the largest model began to encounter performance issues. During CPU inference, constant page faults led to sharp performance declines. For GPU inference, the system experienced device crashes and restarts. These findings highlight that memory is a critical bottleneck for use cases requiring larger models or extended context sizes, posing significant challenges for LLM inference on resource-constrained edge devices. 
Therefore, next-generation edge devices could be equipped with larger memory amounts to be more suitable for memory-intensive GenAI workloads.

\begin{tcolorbox}[
    breakable,
    colback=blue!5!white, 
    colframe=blue!75!black, 
    title=Main Takeaways, 
    fonttitle=\small, 
    fontupper=\small,
    sharp corners, 
    boxrule=0.5pt, 
    left=2pt, right=2pt, top=2pt, bottom=2pt,
    before skip=6pt, after skip=2pt
]
\begin{itemize}
    \item Memory usage grows approximately linearly with model size
    \item GPU inference comes at the cost of significant memory overhead
    \item \qo{} quantization reduces memory footprint for GPU inference but adds substantial memory overhead for CPU inference compared to \qkm{} quantization
    \item Memory is a main bottleneck for using very large models or context sizes on edge devices.
\end{itemize}
\end{tcolorbox}

\subsection{Inference Performance: Latency and Throughput}

\begin{figure}[!th]
    \centering
    \begin{subfigure}{0.32\textwidth}
        \centering
        \includegraphics[width=1\linewidth]{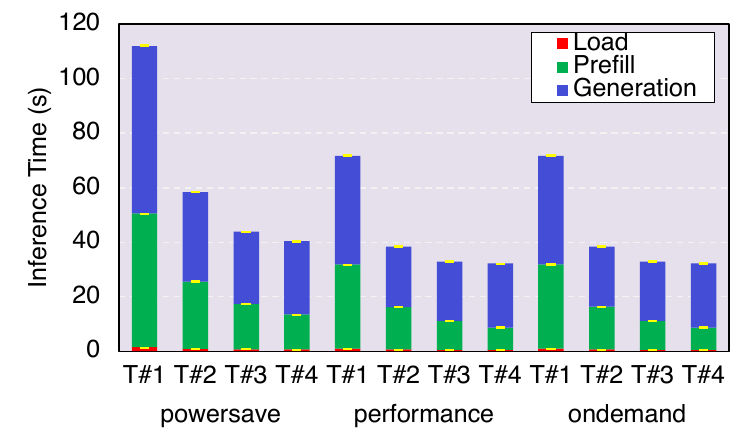}
        \caption{Latency RPi 5, \qkm{}}
        \label{fig:latency_rpi5_q4km}
    \end{subfigure} \hfill
    \begin{subfigure}{0.32\textwidth}
        \centering
        \includegraphics[width=1\linewidth]{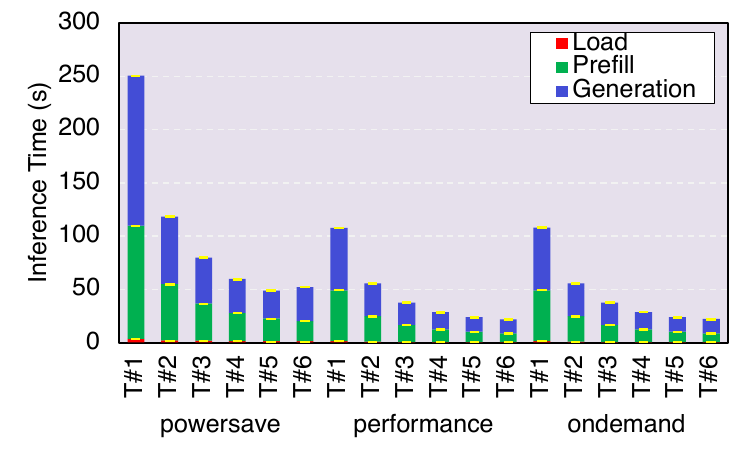}
        \caption{Latency Orin CPU, \qkm{}}
        \label{fig:latency_orin_cpu_q4km}
    \end{subfigure} \hfill
    \begin{subfigure}{0.32\textwidth}
        \centering
        \includegraphics[width=1\linewidth]{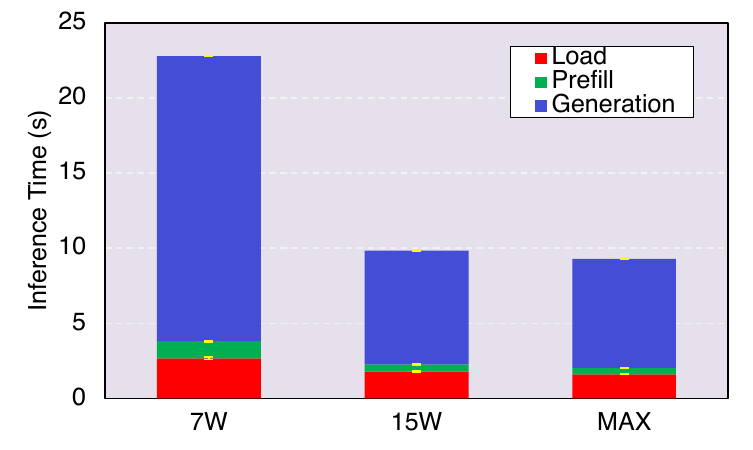}
        \caption{Latency Orin GPU, \qkm{}}
        \label{fig:latency_orin_gpu_q4km}
    \end{subfigure}
    
    \begin{subfigure}{0.32\textwidth}
        \centering
        \includegraphics[width=1\linewidth]{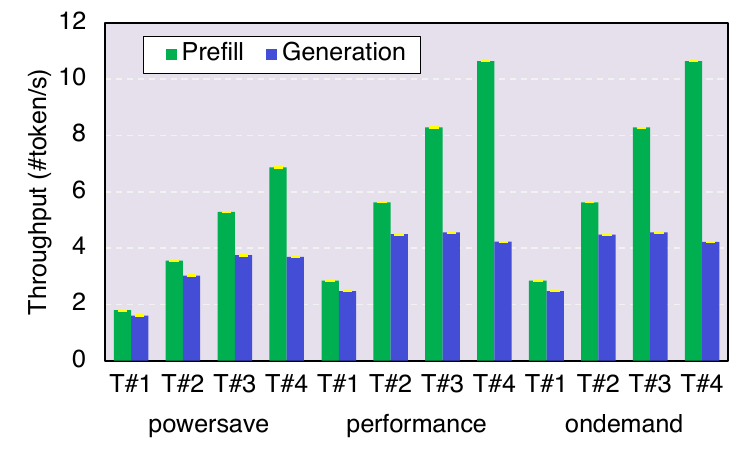}
        \caption{Throughput RPi 5, \qkm{}}
        \label{fig:throughput_rpi5_q4km}
    \end{subfigure} \hfill
    \begin{subfigure}{0.32\textwidth}
        \centering
        \includegraphics[width=1\linewidth]{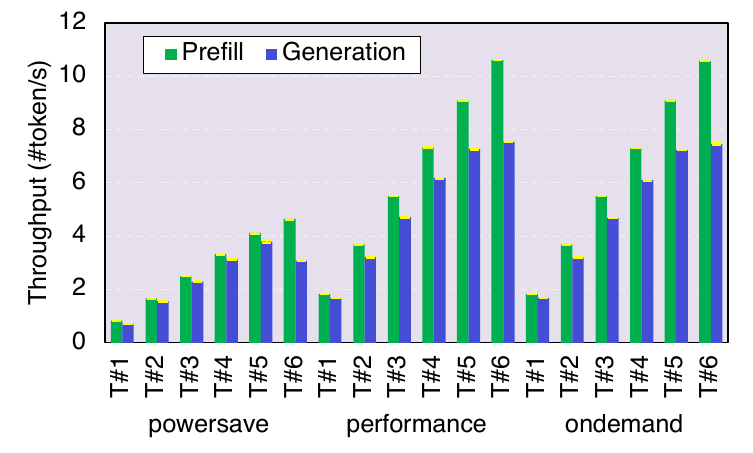}
        \caption{Throughput Orin CPU, \qkm{}}
        \label{fig:throughput_orin_cpu_q4km}
    \end{subfigure} \hfill
    \begin{subfigure}{0.32\textwidth}
        \centering
        \includegraphics[width=1\linewidth]{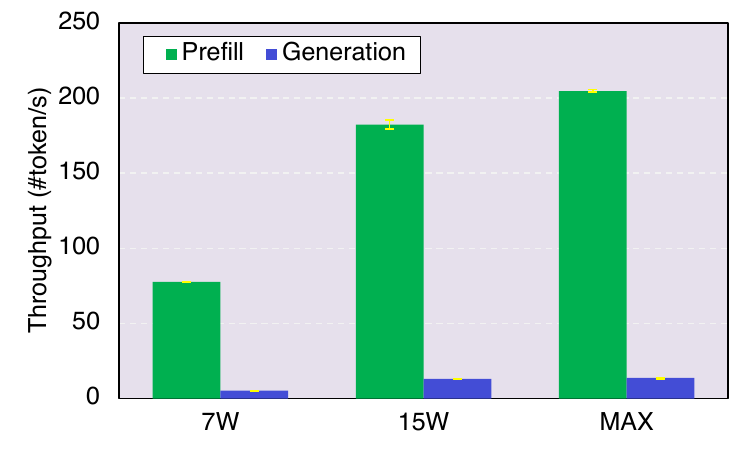}
        \caption{Throughput Orin GPU, \qkm{}}
        \label{fig:throughput_orin_gpu_q4km}
    \end{subfigure}\hfill
    
    \begin{subfigure}{0.32\textwidth}
        \centering
        \includegraphics[width=1\linewidth]{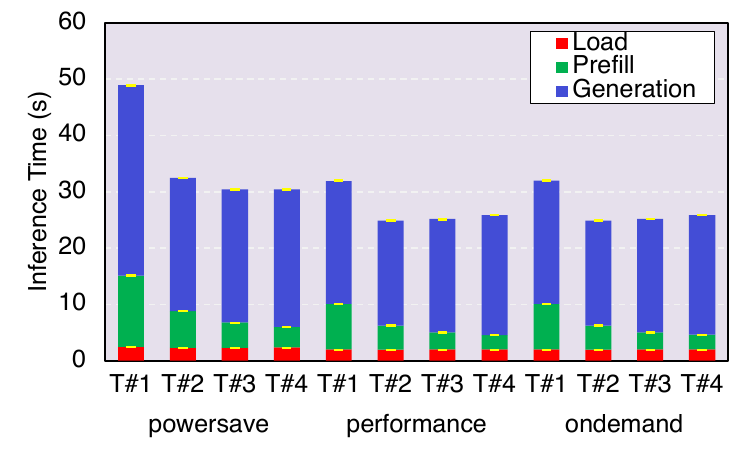}
        \caption{Latency RPi 5, \qo{}}
        \label{fig:latency_rpi5_q40}
    \end{subfigure} \hfill
    \begin{subfigure}{0.32\textwidth}
        \centering
        \includegraphics[width=1\linewidth]{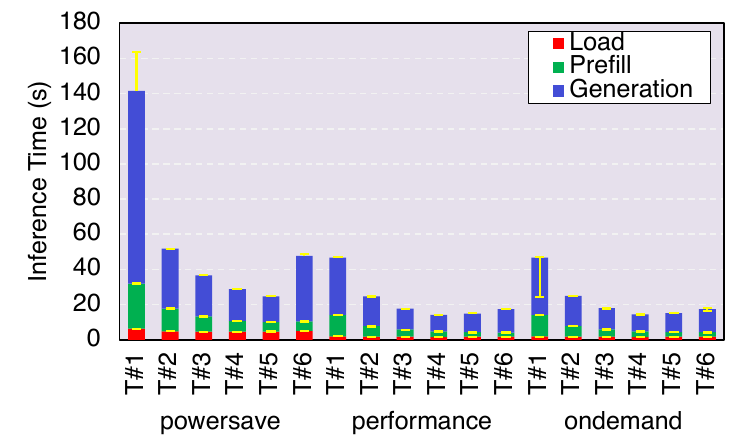}
        \caption{Latency Orin CPU, \qo{}}
        \label{fig:latency_orin_cpu_q40}
    \end{subfigure} \hfill
    \begin{subfigure}{0.32\textwidth}
        \centering
        \includegraphics[width=1\linewidth]{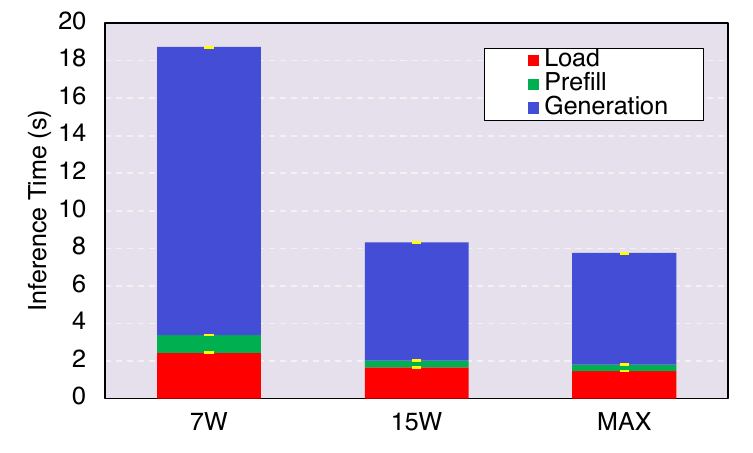}
        \caption{Latency Orin GPU, \qo{}}
        \label{fig:latency_orin_gpu_q40}
    \end{subfigure}\hfill
    
    \begin{subfigure}{0.32\textwidth}
        \centering
        \includegraphics[width=1\linewidth]{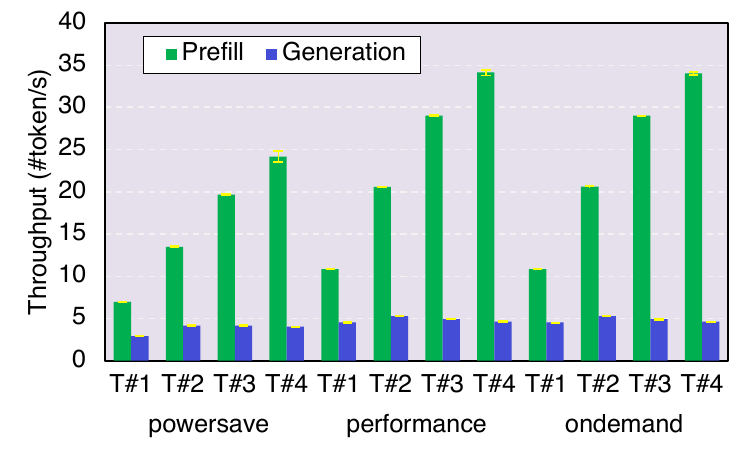}
        \caption{Token throughput RPi 5, \qo{}}
        \label{fig:throughput_rpi5_q40}
    \end{subfigure} \hfill
    \begin{subfigure}{0.32\textwidth}
        \centering
        \includegraphics[width=1\linewidth]{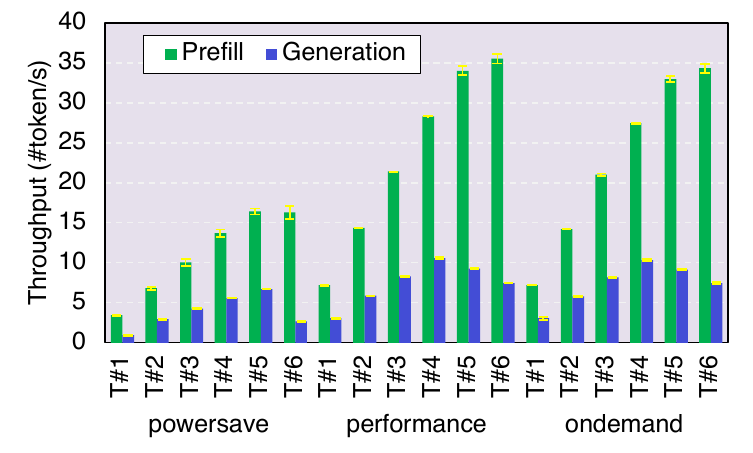}
        \caption{Token throughput Orin CPU, \qo{}}
        \label{fig:throughput_orin_cpu_q40}
    \end{subfigure} \hfill
    \begin{subfigure}{0.32\textwidth}
        \centering
        \includegraphics[width=1\linewidth]{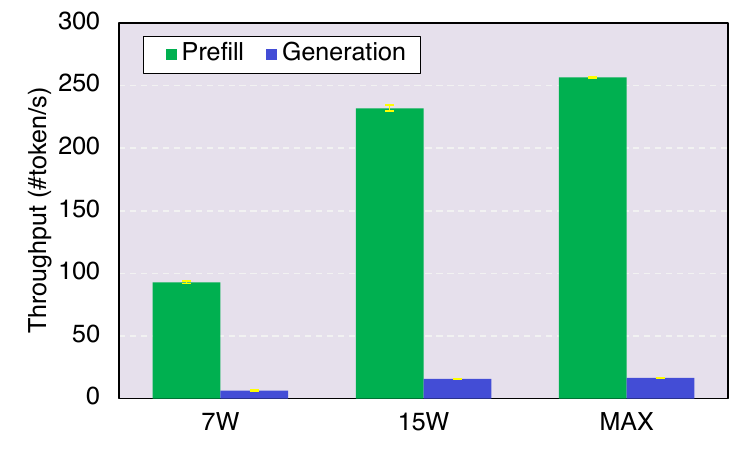}
        \caption{Token throughput Orin GPU, \qo{}}
        \label{fig:throughput_orin_gpu_q40}
    \end{subfigure}\hfill
    
    \caption{Latency and throughput for multiple configurations, all quantization schemes and all devices}
    \label{fig:latency_throughput_overall}
\end{figure}

\begin{figure}[!t]
    \centering
    \begin{subfigure}{0.33\textwidth}
        \centering
        \includegraphics[width=1\linewidth]{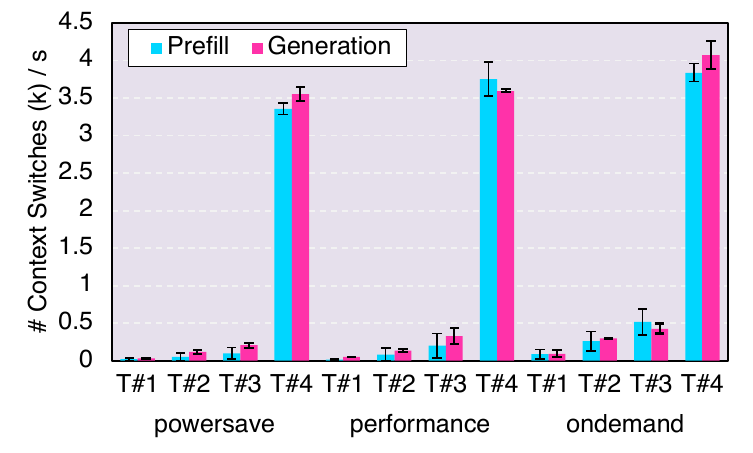}
        \caption{RPi 5, \qkm{}}
        \label{fig:context_switches_rpi5_q4km}
    \end{subfigure}%
    \begin{subfigure}{0.33\textwidth}
        \centering
        \includegraphics[width=1\linewidth]{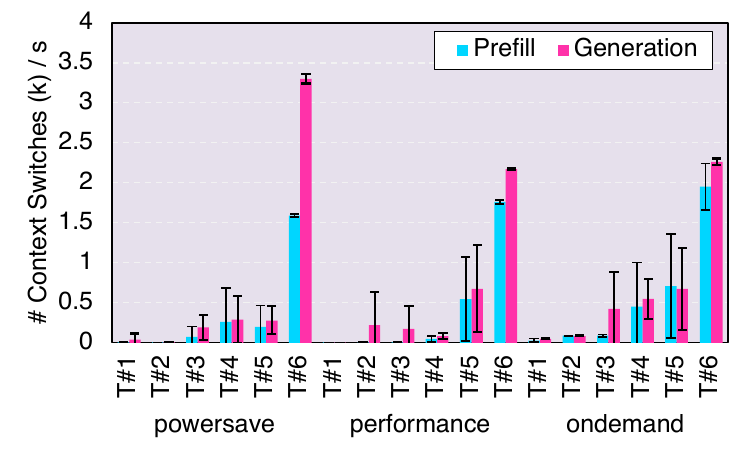}
        \caption{Orin CPU, \qkm{}}
        \label{fig:context_switches_orin_cpu_q4km}
    \end{subfigure}\hfill%
    
    \begin{subfigure}{0.33\textwidth}
        \centering
        \includegraphics[width=1\linewidth]{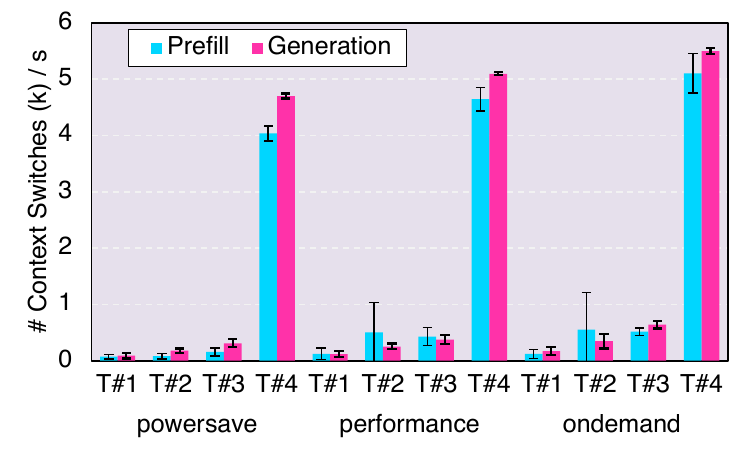}
        \caption{RPi 5, \qo{}}
        \label{fig:context_switches_rpi5_q40}
    \end{subfigure}%
    \begin{subfigure}{0.33\textwidth}
        \centering
        \includegraphics[width=1\linewidth]{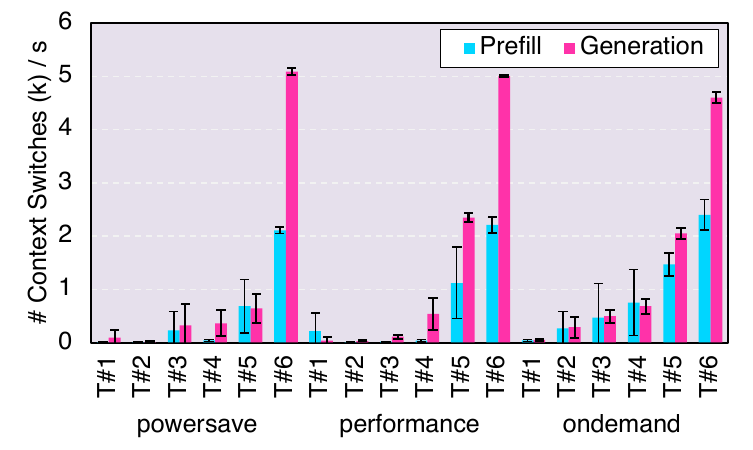}
        \caption{Orin CPU, \qo{}}
        \label{fig:context_switches_orin_cpu_q40}
    \end{subfigure}\hfill%
    
    \caption{Context switches for multiple configuration, all quantization schemes, RPi 5, and Orin CPU}
    \label{fig:context_overall}
\end{figure}

As shown in Figure~\ref{fig:latency_throughput_overall}, end-to-end latency generally decreases with an increasing thread count for both the RPi 5 and the Orin CPU under \qkm{} quantization. However, with \qo{} quantization, the lowest latency was observed at two threads on the RPi 5 and four threads on the Orin CPU when using the \textit{ondemand} and \textit{performance} governors. This reduction was primarily due to a significantly shorter generation phase. Additionally, the \textit{powersave} governor led to higher latencies compared to \textit{ondemand} and \textit{performance} governors. While no significant difference was observed between \textit{ondemand} and \textit{performance} on the RPi 5, the Orin CPU exhibited slightly lower latencies with the \textit{performance} governor across most thread counts. An exception was observed with the \textit{powersave} governor on the Orin, where the six-thread configuration resulted in a performance drop compared to five threads. This behavior was not observed on the RPi 5 but appears to be linked to a significantly higher number of context switches during the generation phase in this configuration, as illustrated in Figure~\ref{fig:context_overall}. On the Orin GPU, the 15W power mode more than halves end-to-end latency compared to 7W mode, with maximum GPU frequency providing further minor latency improvements. When analyzing time to first token (TTFT)—the sum of load latency and prefill latency—distinct patterns emerge based on the quantization scheme. For \qkm{} quantization, CPU inference is dominated by the prefill phase, which accounts for over 90\% of TTFT across all configurations. In contrast, for GPU inference, the load phase is the bottleneck, making up ~80\% of TTFT. This is due to two factors: \emph{(i)} the prefill phase is significantly shorter on the GPU than on the CPU, and \emph{(ii)} he load phase takes 2–2.5 seconds on the GPU, compared to ~1 second for CPU inference. For \qo{} quantization, the CPU inference load phase accounts for a significantly higher portion of TTFT (~43\%) in the fastest configuration. This is due to a substantial increase in loading time, which quadrupled in the worst case compared to \qkm{} quantization, depending on the configuration. Additionally, \qo{} provides drastically lower prefill latencies, which were reduced by a factor of 3 to 4. For GPU inference, \qo{} quantization reduced both load and prefill latencies. However, the reported short load times are only valid for runs 2–5, as the model remains cached in memory between runs. In first-run scenarios, much higher load times were observed, ranging from 5 seconds to over 200 seconds, depending on the model size, quantization scheme, and device. However, the first runs simulate a cold-start scenario, in which the model is only loaded from scratch upon the first arriving request. In a real-world scenario these long load times could be omitted by pre-loading and caching the model in memory.

\begin{figure}[!t]
    \centering
    \begin{subfigure}{0.33\textwidth}
        \centering
        \includegraphics[width=1\linewidth]{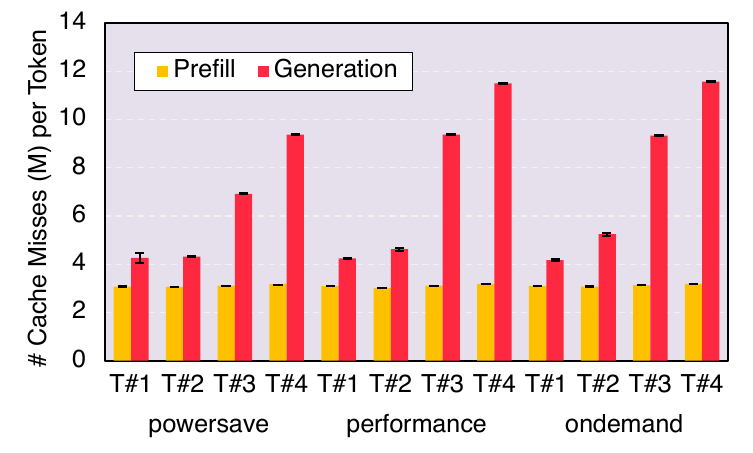}
        \caption{RPi 5, \qkm{}}
        \label{fig:cache_misses_rpi5_q4km}
    \end{subfigure}%
    \begin{subfigure}{0.33\textwidth}
        \centering
        \includegraphics[width=1\linewidth]{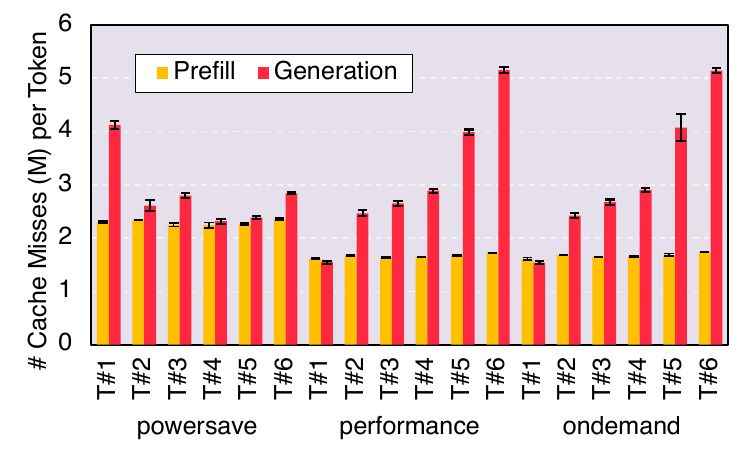}
        \caption{Orin CPU, \qkm{}}
        \label{fig:cache_misses_orin_cpu_q4km}
    \end{subfigure}\hfill%
    
    \begin{subfigure}{0.33\textwidth}
        \centering
        \includegraphics[width=1\linewidth]{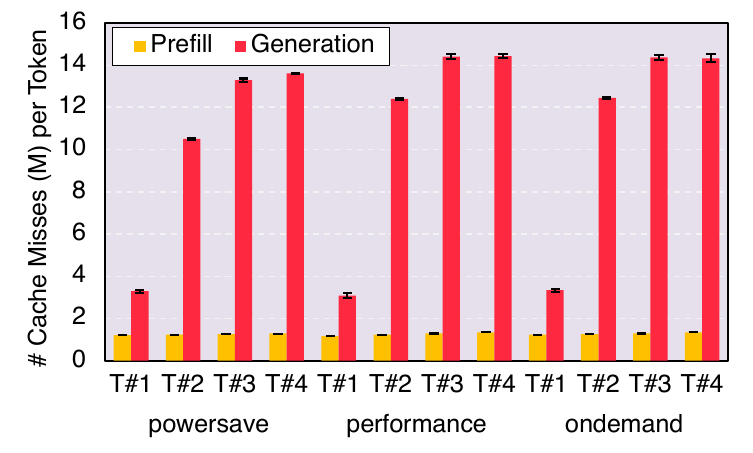}
        \caption{RPi 5, \qo{}}
        \label{fig:cache_misses_rpi5_q40}
    \end{subfigure}%
    \begin{subfigure}{0.33\textwidth}
        \centering
        \includegraphics[width=1\linewidth]{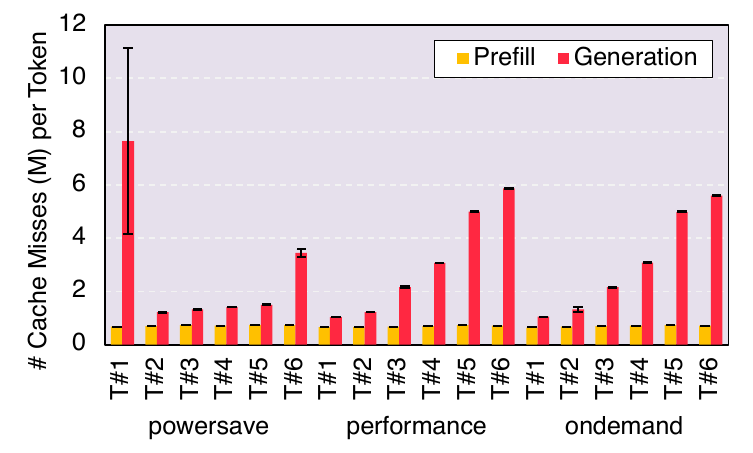}
        \caption{Orin CPU, \qo{}}
        \label{fig:cache_misses_orin_cpu_q40}
    \end{subfigure}\hfill%
    \caption{Level-1-cache misses for multiple configuration, all quantization schemes, RPi 5, and Orin CPU}
    \label{fig:cache_overall}
\end{figure}

As shown in Figure~\ref{fig:latency_throughput_overall}, \qo{} quantization more than tripled prefill throughput and increased generation throughput by $\sim10$--$70\%$ for CPU inference, compared to \qkm{} quantization. This improvement aligns with expectations, as \qo{} quantization incorporates Arm processor optimizations applied during the load phase \cite{optimize_kernels}. Interestingly, GPU inference also exhibited performance gains, varying strongly between 3\% for Llama 3.1 8B and 380\% for Yi 1.5 6B in the prefill phase and between 1\% for Flan T5 Large and 30\% for Gemma 2 9B for the generation phase depending on the model and power mode. The huge prefill improvements of Yi 1.5 6B can be explained by the lower memory usage of the model which eliminates the memory saturation effect already discussed earlier. However, for most setups the performance improvement lay between 20\% and 30 \% in the prefill phase and in the generation phase most smaller models saw a throughput improvement of 5-10\%, while for larger models it typically lay in the range from 15\% to 20\%. 

For CPU inference, prefill phase throughput scales roughly linearly with increasing thread count, with \textit{performance} and \textit{ondemand} governors yielding higher throughputs than the \textit{powersave} governor. Similarly, generation throughput also increases linearly with thread count but eventually plateaus. On the Orin, plateauing occurs at 6 threads while, on the RPi 5, plateauing occurs at 3 or 4 threads with \qkm{} quantization, depending on the CPU governor. This behavior was observed for all models except Flan T5 Large, which did not exhibit a clear performance plateau. For the \textit{powersave} governor, 3 threads yielded the highest throughput, while for \textit{performance} and \textit{ondemand} governors, 2 and 3 threads performed similarly. With \qo{} quantization, 2 and 3 threads also performed similarly under powersave, whereas 2 threads provided the highest throughput for the other two governors. On the Orin CPU with \qo{} quantization, a decline in prefill throughput was observed at 6 threads, while generation throughput peaked at 4 or 5 threads, depending on the model, before decreasing at higher thread counts under the \textit{performance} and \textit{ondemand} governors. For the \textit{powersave} governor, 5 threads provided the best generation throughput. The decline in generation performance at higher thread counts is likely due to the generation phase being memory-bound, where performance is primarily constrained by memory access latency rather than computational power \cite{patel_splitwise_2024}. As a result, increasing compute resources beyond a certain point has only minimal impact on throughput. As observed in our results, increasing the thread number beyond the optimum can even be harmful to performance due to the overhead of saving and restoring the processor state during context switches caused by time-sharing of CPU-cores among the threads. Additionally, background processes competing for CPU-time may displace the llama.cpp threads and cause additional context switches. This effect was observed in our experiments, which is evident from the exploding number of context switches with maximum thread count visible in Figure~\ref{fig:context_overall}. Context switches can, moreover, negatively impact cache performance and, therefore, increase average memory access times, as data used by one thread may replace data cached by the previously running thread and, therefore, cause additional cache misses \cite{mogulEffectContextSwitches1991}. This effect was also found in our experiments for the generation phase as is observable from Figure~\ref{fig:cache_overall}, which explains the degrading performance with larger thread counts beyond a certain point during the generation phase.

For GPU inference, the \textit{15W} power mode improves both prefill by factors of $\sim1.9-2.5$ and generation throughput by factors of $\sim1.3-2.3$ compared to \textit{7W} mode. Using maximum GPU frequencies further enhances prefill throughput by $\sim1-58\%$, and generation throughput by $\sim2-12\%$ improvements. Generally, a trend can be identified that the \textit{MAX}-mode yields more pronounced performance improvements for smaller models than for larger models. For two configurations --- Llama 3.1 8B with \qo{} quantization and Gemma 2 9B with \qkm{} quantization --- even small performance drops in prefill throughput were observable.

\begin{figure}[!t]
    \centering
    \begin{subfigure}{0.32\textwidth}
        \centering
        \includegraphics[width=1\linewidth]{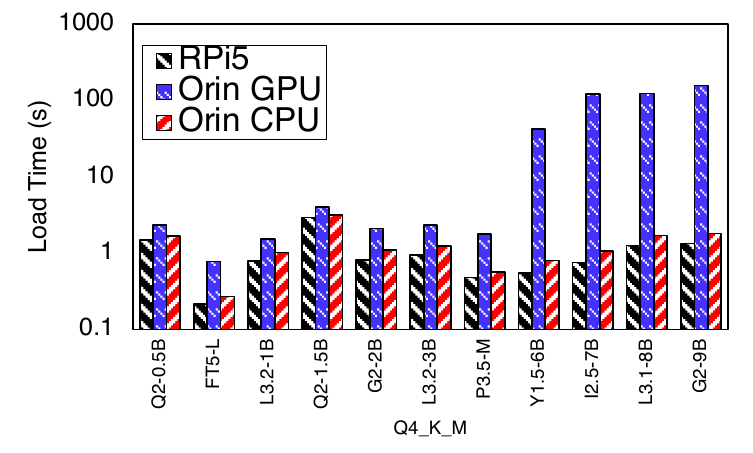}
        \caption{Load time, Q4\_K\_M}
        \label{fig:load_time_models_k}
    \end{subfigure}
    \begin{subfigure}{0.32\textwidth}
        \centering
        \includegraphics[width=1\linewidth]{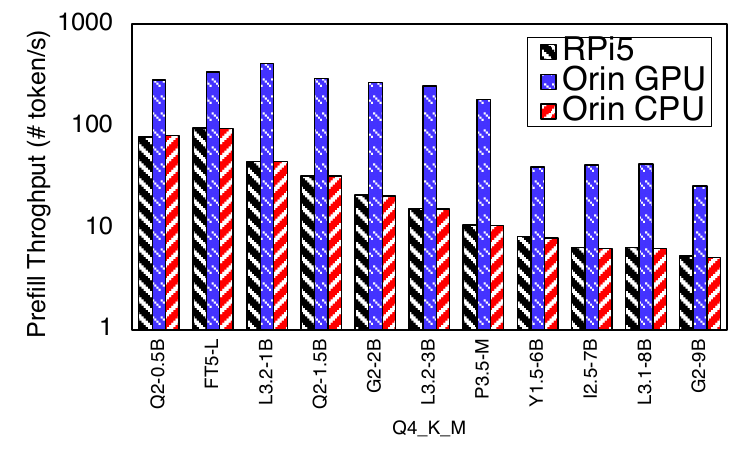}
        \caption{Prefill Throughput, Q4\_K\_M}
        \label{fig:prefill_throughput_models_k}
    \end{subfigure} \hfill
    \begin{subfigure}{0.32\textwidth}
        \centering
        \includegraphics[width=1\linewidth]{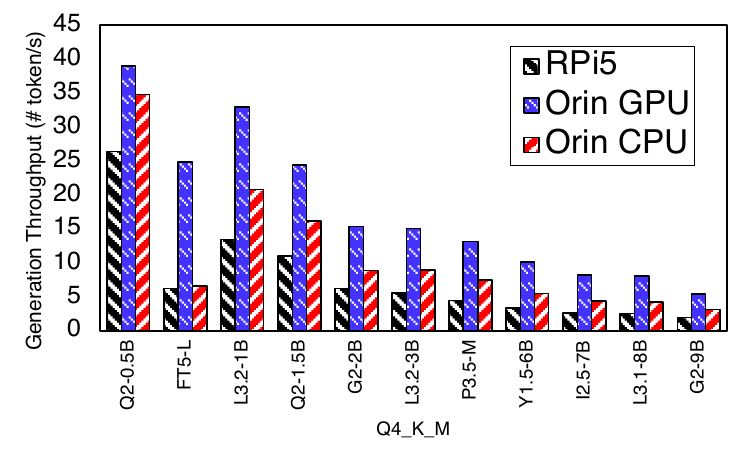}
        \caption{Generation Throughput, Q4\_K\_M}
        \label{fig:gen_throughput_models_k}
    \end{subfigure} \hfill

    \begin{subfigure}{0.32\textwidth}
        \centering
                \includegraphics[width=1\linewidth]{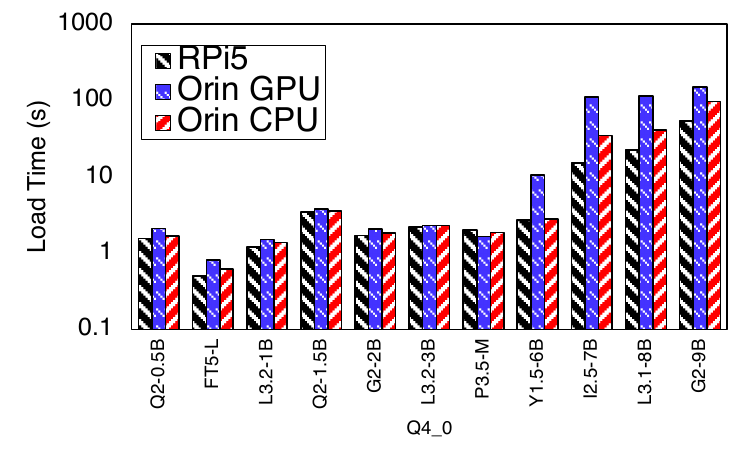}
        \caption{Load Time, Q4\_0}
        \label{fig:load_time_models_0}
    \end{subfigure}
    \begin{subfigure}{0.32\textwidth}
        \centering
                \includegraphics[width=1\linewidth]{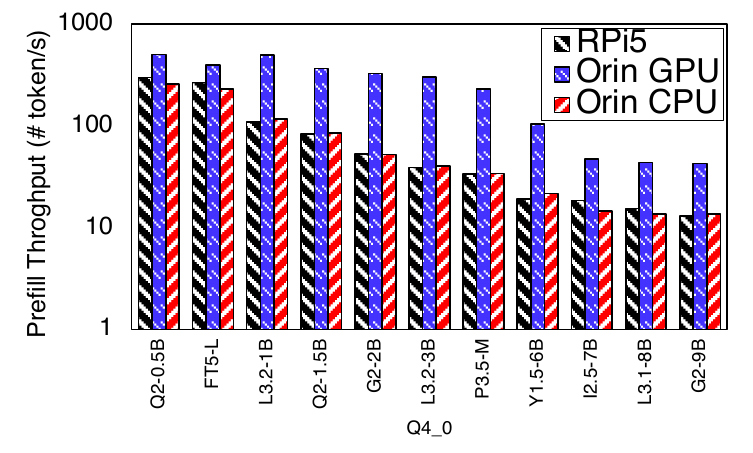}
        \caption{Prefill Throughput, Q4\_0}
        \label{fig:prefill_throughput_models_0}
    \end{subfigure} \hfill
    \begin{subfigure}{0.32\textwidth}
        \centering
            \includegraphics[width=1\linewidth]{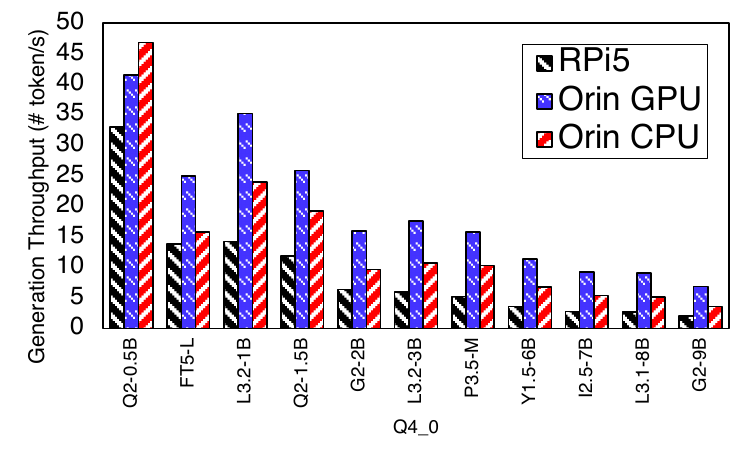}
        \caption{Generation Throughput, Q4\_0}
        \label{fig:gen_throughput_models_0}
    \end{subfigure} \hfill
  
    \caption{Load time and throughput for all models, quantization schemes, and all devices}
    \label{fig:latency_throughput_models_overall}
\end{figure}

Looking at Figure~\ref{fig:latency_throughput_models_overall}, it is evident that load time is higher for GPU inference with \qkm{} quantization. However, with \qo{} quantization, GPU load time is shorter for most models up to Phi 3.5 Mini, as CPU loading times increase by factors ranging from $\sim1.05$× for Qwen 2 0.5B to $\sim4$× for Phi 3.5 Mini. For larger models, such as Yi 1.5 6B and beyond, load times increase dramatically, with factors between $\sim17$× and $\sim53$× compared to \qkm{} quantization. This occurs because the high memory footprint forces models to be evicted from memory during runs, requiring them to be reloaded from storage for each inference, significantly degrading performance. Additionally, a higher number of page faults was observed for these larger models, further increasing load times due to the slow nature of storage accesses. For \qkm{} quantization, this issue was only noticeable in GPU inference. However, for \qo{} quantization, it also affected CPU inference, which can be explained by the increased memory footprint, as discussed in Subsection~\ref{sub:mem}.

Figure~\ref{fig:latency_throughput_models_overall} reveals that GPU inference consistently delivers the highest prefill throughput across all models and quantization schemes. In contrast, CPU inference shows comparable performance between the RPi 5 and Orin, except for Qwen 2 0.5B and Flan T5 Large under \qo{} quantization, where the RPi 5 achieves noticeably better results. A general trend of decreasing prefill throughput with increasing model size can be observed. However, Llama 3.2 1B stands out as an exception, significantly exceeding expectations in GPU inference despite its relatively small size. Interestingly, this advantage was not present in CPU inference, contradicting Meta’s claim that Llama 3.2 models are optimized for Arm processors \cite{llama32_annoucement}. Additionally, the largest models suffer a sharp drop in prefill performance, a consequence of their high memory footprint, as discussed earlier.

GPU inference achieves the highest generation throughput across all models, except for Qwen 2 0.5B under \qo{} quantization. However, the performance gap between CPU and GPU inference is much smaller than in the prefill phase. This can be attributed to the memory-bound nature of the generation phase, where increased computational power has less impact on performance improvements. For CPU inference, the Orin significantly outperforms the RPi 5, with throughput gains of $~32-67\%$ for \qkm{} quantization and $\sim43-94\%$ for \qo{} quantization, depending on the model. However, for Flan T5 Large, this advantage is much less pronounced, with only $\sim5\%$ and $\sim14\%$ increases, respectively. This behavior can be explained by the Orin’s larger cache size, which reduces cache misses, as seen in Figure~\ref{fig:cache_overall}. Due to the memory boundness of the generation phase a lower number of cache misses and, hence, faster memory access times have a significant impact on performance. Based on this result we believe, that there is room for optimizing the memory hierarchy and optimizing memory accesses for LM inference on future-generation edge devices to achieve faster token generation speeds.

When comparing model performance, the general trend of decreasing throughput with increasing model size remains evident. Notable exceptions include Flan T5 Large and Gemma 2 models, which underperform relative to their size expectations. Unlike in the prefill phase, no major performance drops were observed for the largest models during generation.

\begin{tcolorbox}[
    breakable,
    colback=blue!5!white, 
    colframe=blue!75!black, 
    title=Main Takeaways, 
    fonttitle=\small, 
    fontupper=\small,
    sharp corners, 
    boxrule=0.5pt, 
    left=2pt, right=2pt, top=2pt, bottom=2pt,
    before skip=6pt, after skip=2pt
]
\begin{itemize}
    \item GPU inference outperforms CPU inference in terms of token throughput at the cost of higher load times
    \item \qo{} quantization substantially improves token throughput for all devices but induces higher load times for CPU inference than \qkm{} quantization
    \item Token throughput follows approximately a hyperbolic relationship with model size
\end{itemize}
\end{tcolorbox}

\subsection{Energy Consumption and Efficiency Trade-Offs}

\begin{figure}[!th]
    \centering
    \begin{subfigure}{0.32\textwidth}
        \centering
                \includegraphics[width=1\linewidth]{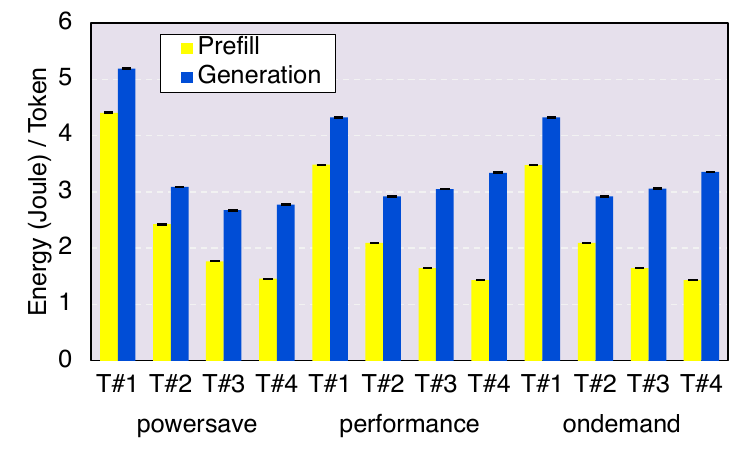}
        \caption{RPi 5, \qkm{}}
        \label{fig:energy_rpi5_km}
    \end{subfigure} \hfill
    \begin{subfigure}{0.32\textwidth}
        \centering
                        \includegraphics[width=1\linewidth]{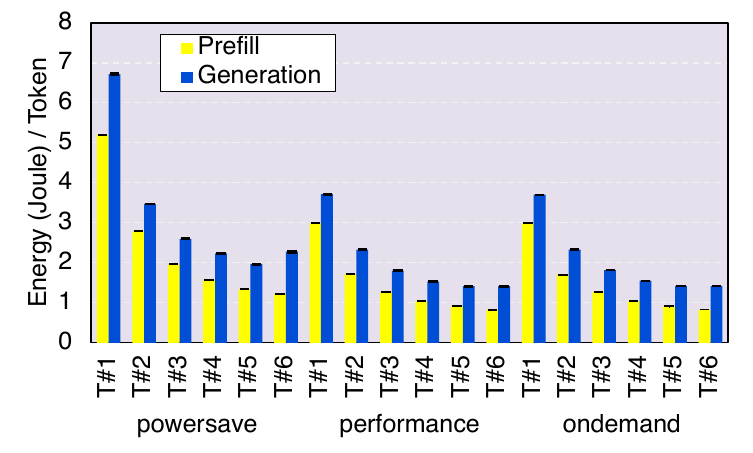}
        \caption{Orin CPU, \qkm{}}
        \label{fig:energy_orin_cpu_km}
    \end{subfigure} \hfill
    \begin{subfigure}{0.32\textwidth}
        \centering
                        \includegraphics[width=1\linewidth]{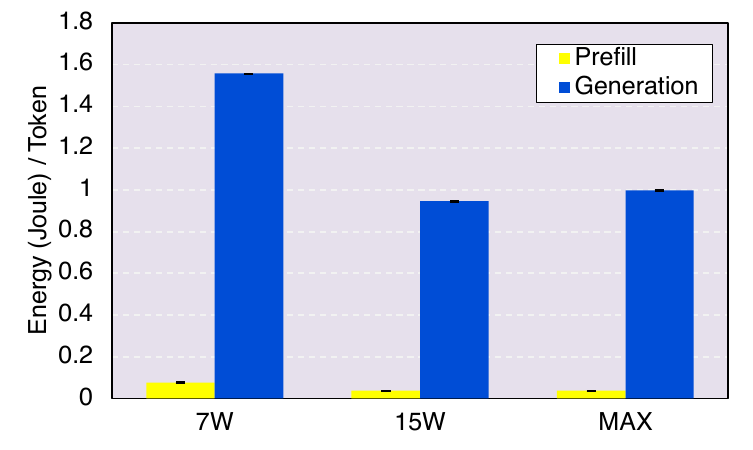}
        \caption{Orin GPU, \qkm{}}
        \label{fig:energy_orin_gpu_km}
    \end{subfigure}

    \begin{subfigure}{0.32\textwidth}
        \centering
                        \includegraphics[width=1\linewidth]{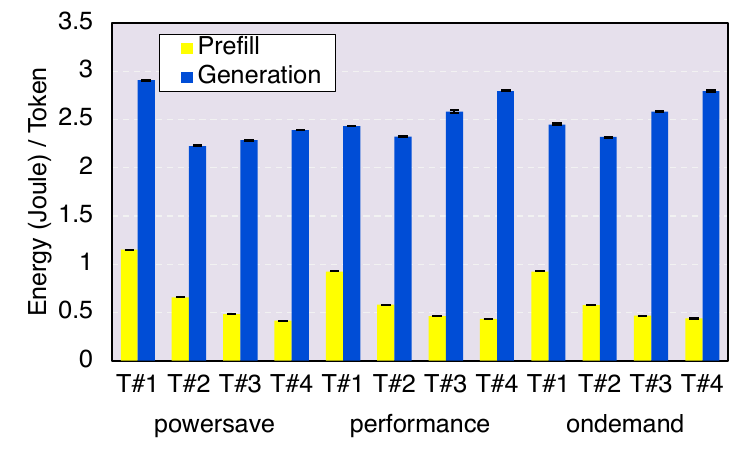}
        \caption{RPi 5, \qo{}}
        \label{fig:energy_rpi5_0}
    \end{subfigure} \hfill
    \begin{subfigure}{0.32\textwidth}
        \centering
                        \includegraphics[width=1\linewidth]{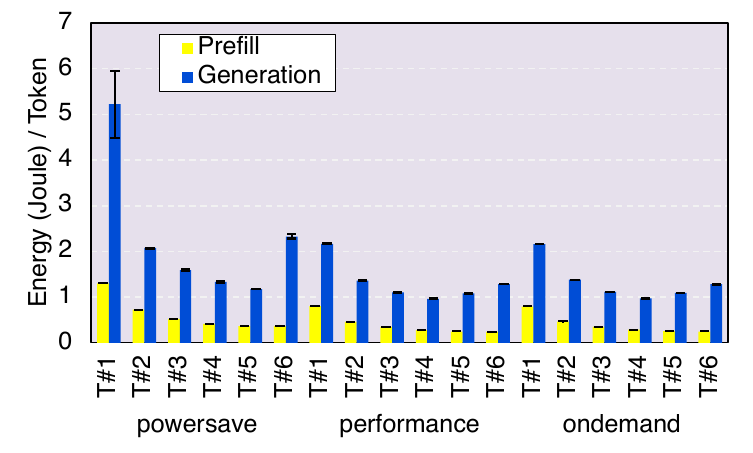}
        \caption{Orin CPU, \qo{}}
        \label{fig:energy_orin_cpu_0}
    \end{subfigure} \hfill
    \begin{subfigure}{0.32\textwidth}
        \centering
                        \includegraphics[width=1\linewidth]{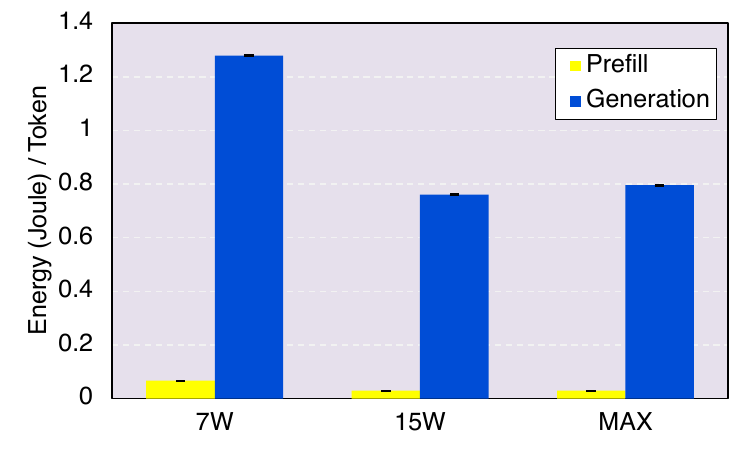}
        \caption{Orin GPU, \qo{}}
        \label{fig:energy_orin_gpu_0}
    \end{subfigure}
    
    \caption{Energy consumption per token for all configurations, quantization schemes, and devices}
    \label{fig:energy_overall}
\end{figure}

In this subsection, we analyze the energy efficiency of LM inference across different configurations. Energy consumption is a critical factor in edge systems, particularly for battery-powered devices, where optimizing power usage directly impacts device longevity and real-world feasibility. Figure~\ref{fig:energy_overall} illustrates the energy consumption per token for the Phi 3.5 Mini model across all tested configurations and devices. In most cases, the fastest configurations also exhibit the lowest energy consumption per token for both prefill and generation phases, as the reduction in inference time outweighs the increased power draw. However, some notable exceptions were observed. On the RPi 5, the \textit{powersave} governor with optimal thread count resulted in lower energy consumption per token than the \textit{ondemand} and \textit{performance} governors due to its reduced power draw, despite the longer inference time. On the Orin GPU, the \textit{15W} mode consumed up to 7\% less energy per token than the \textit{MAX} mode during generation phase, as the power savings compensated for the higher inference time. In contrast, the \textit{MAX} mode resulted in in up to 25\% energy consumption per token for all models up to Yi 1.5 6B during the prefill phase compared to the \textit{15W} mode with the difference being generally more pronounced for smaller models. On the RPi 5, the optimal configuration in the prefill phase was 4 threads across all models, governors, and quantization schemes. However, in the generation phase, the optimal configuration varied depending on the model and quantization scheme. Flan T5 Large was a clear outlier, where 4 threads minimized energy consumption for all governors and quantization schemes. For all other models, under \qkm{} quantization, 2 threads provided the best energy efficiency for both the \textit{performance} and \textit{ondemand} governors, while the \textit{powersave} governor performed best with 3 threads, with the only exception being Qwen 2 0.5B, for which 4 threads with the \textit{powersave} governor and 3 threads with the other two governors minimized energy consumption.
In the case of \qo{} quantization, 2 threads were optimal for all models under the \textit{performance} and \textit{ondemand} governors, while for the \textit{powersave} governor either 3 or 2 threads were optimal depending on the model. 

On the Orin CPU, the \textit{powersave} governor generally exhibited higher energy consumption than the \textit{performance} and \textit{ondemand} governors, which showed no significant differences in energy usage. In the prefill phase, the maximum thread count of 6 proved to be optimal across all models, governors, and quantization schemes. For the generation phase, under \qkm{} quantization, 5 threads yielded the best energy efficiency across all models and governors, except for Flan T5 Large with the \textit{performance} and \textit{powersave} governors, where 6 threads resulted in minimal energy consumption per token for the \textit{ondemand} and \textit{performance} governors. When considering the \qo{} quantization, 4 threads were optimal for the \textit{performance} and \textit{ondemand} governors, while 5 threads were best for the \textit{powersave} governor across all models. The only exception was Flan T5 Large, which also showed optimal energy consumption with 5 threads under the \textit{performance} and \textit{ondemand} governors.

A comparison of quantization schemes reveals that \qo{} quantization significantly reduces energy consumption. In the prefill phase, energy usage is lowered by factors of $\sim2.5$ to $\sim5$ for CPU inference. In the generation phase, CPU energy consumption decreases by $\sim10$–$\sim70\%$, depending on the device and configuration. For GPU inference, energy consumption is reduced by $\sim10$–$\sim70\%$ during prefill phase and up to 20\% during generation phase. This reduction is not solely due to increased inference speed but also to a moderate decrease in power draw compared to \qkm{} quantization.

\begin{figure}[!t]
    \centering
    \begin{subfigure}{0.33\textwidth}
        \centering
                \includegraphics[width=\linewidth]{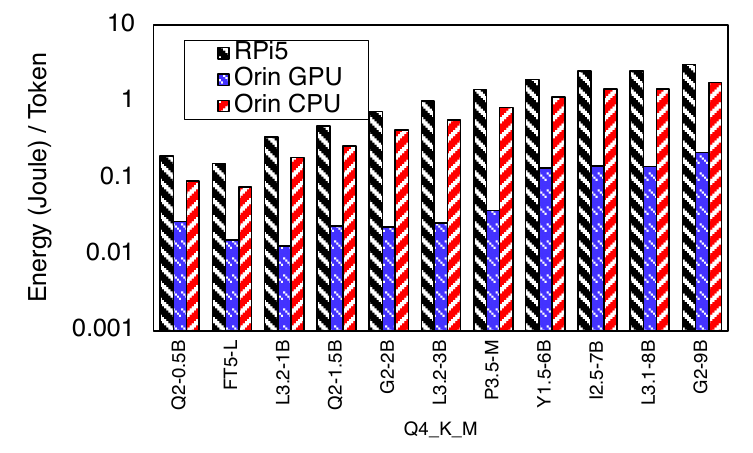}
        \caption{Prefill phase, \qkm{}}
        \label{fig:prefill_energy_models_k}
    \end{subfigure}
    \begin{subfigure}{0.33\textwidth}
        \centering
                        \includegraphics[width=\linewidth]{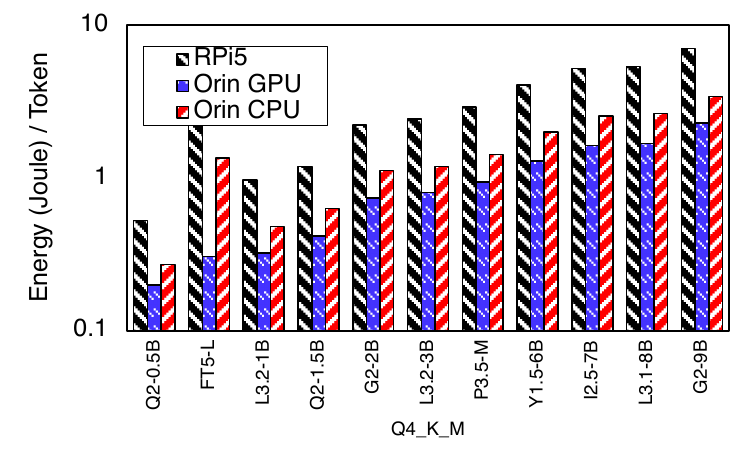}
        \caption{Gen. phase, \qkm{}}
        \label{fig:gen_energy_models_k}
    \end{subfigure} \hfill
    
    \begin{subfigure}{0.33\textwidth}
        \centering
                        \includegraphics[width=\linewidth]{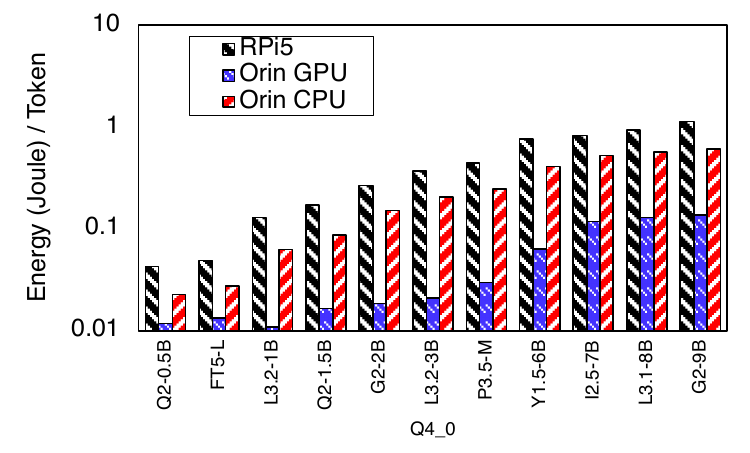}
        \caption{Prefill phase, \qo{}}
        \label{fig:prefill_energy_models_0}
    \end{subfigure}
    \begin{subfigure}{0.33\textwidth}
        \centering
                        \includegraphics[width=\linewidth]{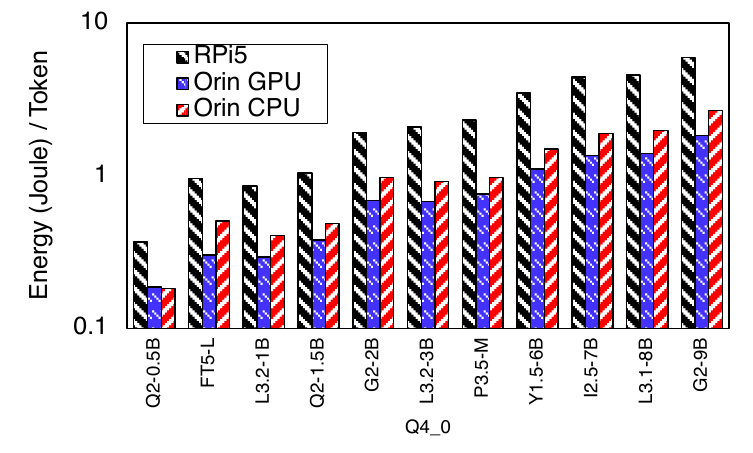}
        \caption{Gen. phase, \qo{}}
        \label{fig:gen_energy_models_0}
    \end{subfigure} \hfill
  
    \caption{Energy per token for all models, quantization schemes, and all devices}
    \label{fig:energy_models_overall}
\end{figure}

When examining model-specific energy consumption in Figure~\ref{fig:energy_models_overall}, a general trend of increasing energy usage with model size is evident. This is primarily driven by longer inference latencies, although a slight increase in power draw is also observed. However, for Phi 3.5 and all larger models, average power draw remained approximately constant. In the generation phase, Flan T5 Large emerged as a notable outlier, exhibiting substantially higher energy consumption relative to its size. This inefficiency stems from its lower generation speed, leading to longer active power draw.

For all models, GPU inference significantly reduced energy consumption compared to the RPi 5, with improvements by factors of $\sim3.6$–$\sim39.3$ in the prefill phase and $\sim1.9$–$\sim7.8$ in the generation phase, depending on the model and quantization scheme. For CPU inference, the Orin also demonstrated superior energy efficiency over the RPi 5 in both phases. In the prefill phase, this improvement was primarily due to a lower mean power draw, resulting in energy reductions of $\sim36.0$–$\sim52.5\%$. In the generation phase, the lower inference latencies on the Orin further amplified this difference, leading to energy consumption reductions of $\sim47$–$\sim58\%$. \qo{} quantization consistently outperformed \qkm{} quantization in terms of energy efficiency for almost all models and devices, mainly due to reduced inference time but also because of a slightly lower power draw. In the prefill phase, the efficiency improvement factors ranged between $\sim1.1$ and $\sim4.5$, depending on the model and device. In the generation phase, improvements were $\sim1.008$ to $\sim2.6$.

\begin{tcolorbox}[
    breakable,
    colback=blue!5!white, 
    colframe=blue!75!black, 
    title=Main Takeaways, 
    fonttitle=\small, 
    fontupper=\small,
    sharp corners, 
    boxrule=0.5pt, 
    left=2pt, right=2pt, top=2pt, bottom=2pt,
    before skip=6pt, after skip=2pt
]
\begin{itemize}
    \item GPU inference is more energy efficient than CPU inference in both prefill and generation phase
    \item \qo{} quantization improves energy efficiency on all devices for both prefill and generation phase over \qkm{} quantization
\end{itemize}
\end{tcolorbox}

\subsection{Quantization Effect}

\begin{figure}[!th]
    \centering
    \begin{subfigure}{\textwidth}
        \centering
        \includegraphics[width=1\linewidth]{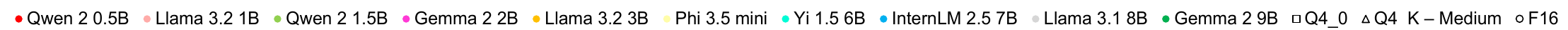}
    \end{subfigure} \hfill
    
    \begin{subfigure}{0.48\textwidth}
        \centering
        \includegraphics[width=1\linewidth]{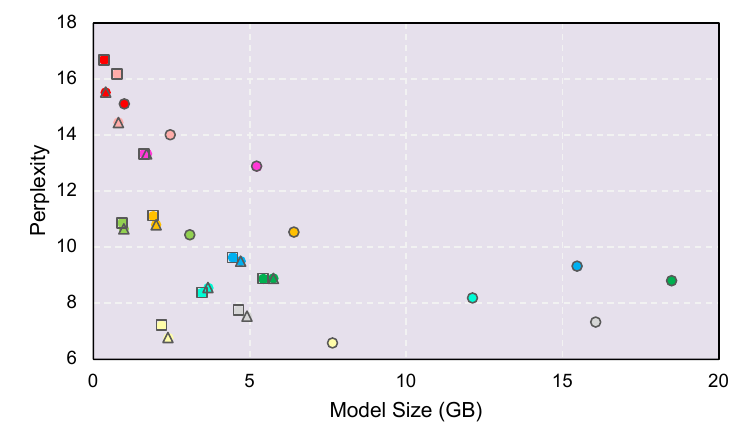}
        \caption{perplexity}
    \end{subfigure}
    \begin{subfigure}{0.48\textwidth}
        \centering
        \includegraphics[width=1\linewidth]{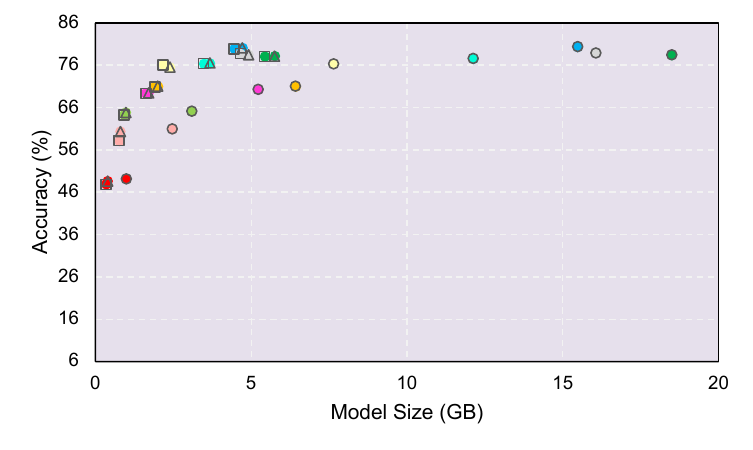}
        \caption{HellaSwag}
    \end{subfigure}\hfill
    
    \begin{subfigure}{0.48\textwidth}
        \centering
        \includegraphics[width=1\linewidth]{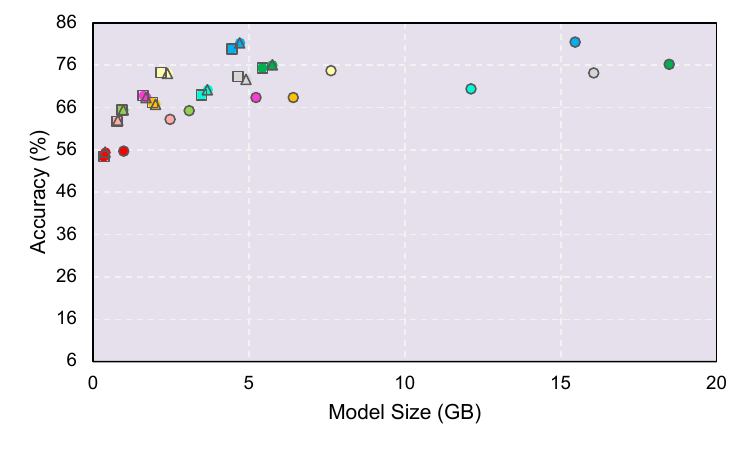}
        \caption{Winogrande}
    \end{subfigure}
    \begin{subfigure}{0.48\textwidth}
        \centering
        \includegraphics[width=1\linewidth]{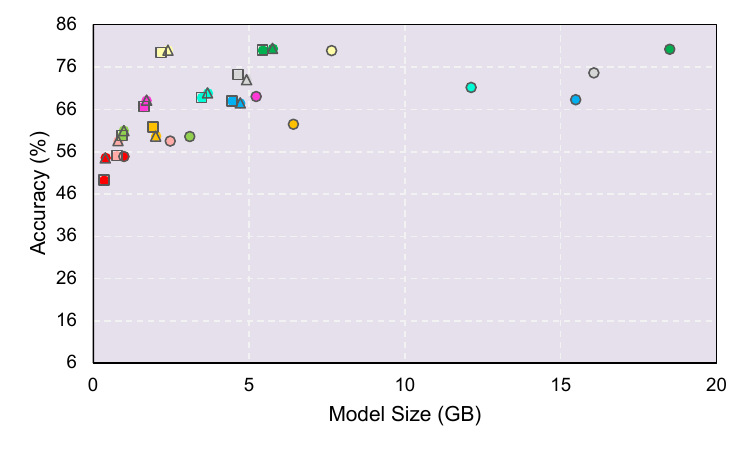}
        \caption{ARC easy}
    \end{subfigure}\hfill
    
    \begin{subfigure}{0.48\textwidth}
        \centering
        \includegraphics[width=1\linewidth]{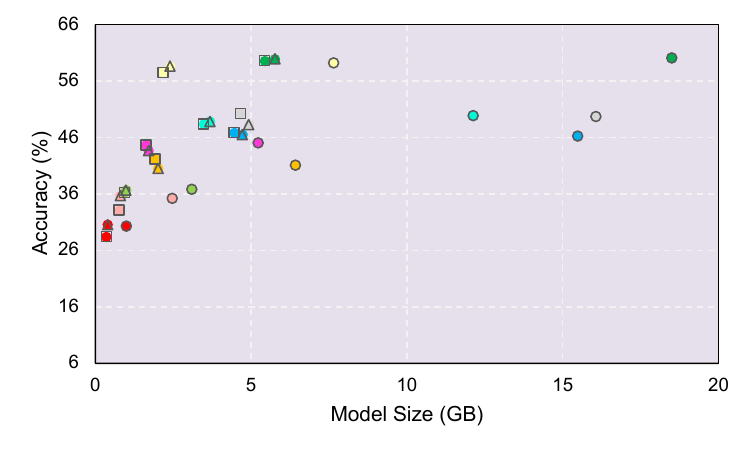}
        \caption{ARC challenge}
    \end{subfigure} 
    \begin{subfigure}{0.48\textwidth}
        \centering
        \includegraphics[width=1\linewidth]{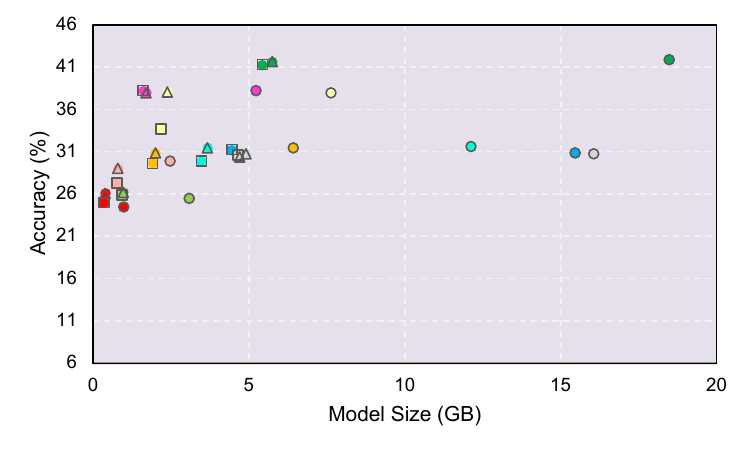}
        \caption{TruthfulQA}
    \end{subfigure} \hfill
    \caption{Results of qualitative benchmarks}
    \label{fig:quality_overall}
\end{figure}

Since quantized models were used in this evaluation, it is essential to assess how quantization impacts both output quality and computational performance compared to unquantized models. To analyze the effects on output quality, a a suite of five benchmarks were conducted, with results presented in Figure~\ref{fig:quality_overall}. The findings reveal a perplexity increase across both quantization schemes and all models when compared to their 16-bit floating point counterparts. This effect was particularly pronounced for the smallest models. Additionally, a clear distinction between quantization schemes emerged: \qo{} quantization exhibited higher perplexity than \qkm{} quantization for all models, except for Yi 1.5 6B and the Gemma models, where both schemes performed similarly. However, on the down-stream task benchmarks the observed accuracy drop was almost in all cases negligible, showing the effectiveness of quantization as a model compression technique. These results go hand in hand with previous work which has shown that 4-bit quantization of LMs impacts model quality on downstream tasks only modestly, while perplexity was found to be a more sensitive metric for evaluating quantization effects \cite{dettmers_4bit_2023}.

\begin{figure}[!th]
    \centering
    \begin{subfigure}{0.99\textwidth}
        \centering
        \includegraphics[width=\linewidth]{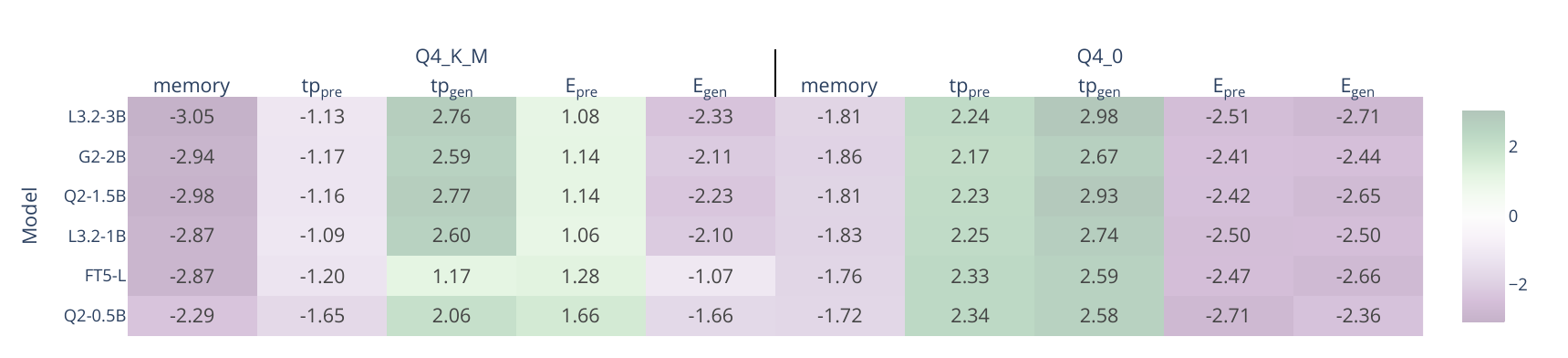}
        \caption{}
        \label{fig:quant_pi5}
    \end{subfigure}
    \begin{subfigure}{0.99\textwidth}
        \centering
        \includegraphics[width=\linewidth]{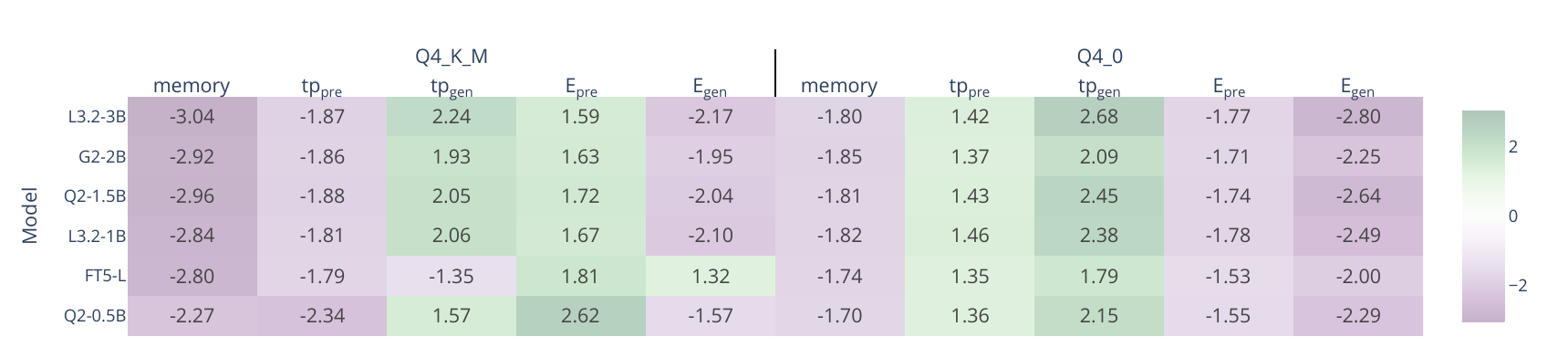}
        \caption{}
        \label{fig:quant_orin_cpu}
    \end{subfigure}
    \begin{subfigure}{0.99\textwidth}
        \centering
        \includegraphics[width=\linewidth]{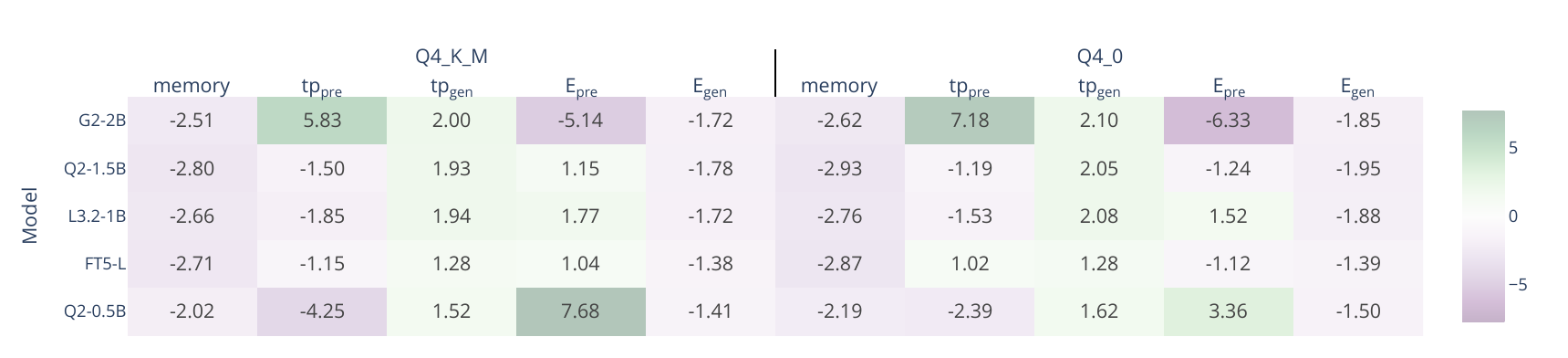}
        \caption{}
        \label{fig:quant_gpu}
    \end{subfigure}
    \caption{Change factors on various metrics on RPi 5 (a), Orin CPU (b), and Orin GPU (c) compared to 16 bit floating point models. Negative numbers indicate decrease by absolute value.}
    \label{fig:quant_all}
\end{figure}

To evaluate the impact of quantization on computational performance, Figure~\ref{fig:quant_all} illustrate how quantization affected memory footprint, token throughput, and energy consumption for all models that were small enough to fit their full-precision versions into memory. In terms of memory footprint reduction, as shown in Figure~\ref{fig:quant_all}, quantization significantly reduces memory usage, with reductions by factors between 1.7 and 3 across all models, devices, and quantization schemes. However, as previously discussed, \qo{} quantization exhibits a higher memory footprint for CPU inference but a lower footprint for GPU inference compared to \qkm{} quantization. Analyzing the effects on token throughput and energy consumption, \qkm{} quantization negatively impacted both token throughput and energy consumption in the prefill phase, with varying intensity depending on the model and device. Throughput reductions ranged between factors of 1.09 and 4.25, while energy consumption increased by factors between 1.04 and 7.57. Conversely, in the generation phase, \qkm{} quantization led to performance improvements, where token throughput increased by factors between 1.17 and 2.76, and energy consumption decreased by factors ranging from 1.02 to 2.17. When it comes to outliers, Gemma 2 2B exhibited unexpected improvements in both prefill throughput and energy consumption for GPU inference. Flan T5 Large showed degraded performance in generation throughput and energy consumption on the Orin CPU. In contrast, \qo{} quantization yielded consistent improvements in both throughput and energy consumption across both phases for CPU inference, while varying results were observed for GPU inference depending on the model. For Gemma 2 2B  prefill throughput improved significantly by a factor of 7.18 and energy consumption per token dropped by a factor of 6.33. For other models prefill performance either dropped or stayed approximately constant compared to unquantized models. However, for Qwen 2 1.5B and Flan T5 Large energy consumption dropped by factors 1.24 and 1.12 respectively, while for Llama 3.2 1B and Qwen 2 0.5B energy consumption increased by factors of 1.52 and 3.36. The improved performance in the generation phase can be attributed to the memory-bound nature of this phase. Since quantized model weights require lower precision, a greater portion of them fits into cache, leading to fewer cache misses and ultimately faster token generation.

\begin{tcolorbox}[
    breakable,
    colback=blue!5!white, 
    colframe=blue!75!black, 
    title=Main Takeaways, 
    fonttitle=\small, 
    fontupper=\small,
    sharp corners, 
    boxrule=0.5pt, 
    left=2pt, right=2pt, top=2pt, bottom=2pt,
    before skip=6pt, after skip=2pt
] 
\begin{itemize}
    \item In terms of perplexity, \qkm{} quantization performed better than \qo{} quantization for most models, especially for the smallest evaluated models. Overall, perplexity increases was especially pronounced for the smallest models.
    \item \qo{} quantization improved memory footprint, token throughput, and energy consumption over full-precision models, while \qkm{} quantization resulted in inferior throughput and energy consumption in the prefill phase.
\end{itemize}
\end{tcolorbox}

\section{Analysis and Trade-Offs Discussion}
\label{sec:discuss}

Optimizing LM inference at the edge requires balancing multiple constraints, including computational performance, energy efficiency, and cost. In this section, we leverage our empirical measurements to analyze key trade-offs, assess its practical usability in real-world deployments, and evaluate the economic feasibility of LM inference on edge devices.

\subsection{Throughput-Energy Trade-Off}

On the Orin CPU, the fastest configuration was either also the most energy-efficient or, at worst, only marginally less efficient than the optimal configuration. However, no single configuration simultaneously maximized performance and energy efficiency on the RPi 5 or the Orin GPU. To determine the optimal trade-off between token throughput and energy consumption, we define the metric action per token $S_\mathrm{t}$ as the ratio between the energy $E$ consumed during an inference phase and the token throughput $t$:

\[S_\mathrm{t}=\frac{E}{t}\]

This metric represents the amount of action required to process a token during the prefill phase or generate a token in the generation phase. A lower action per token value indicates higher efficiency and is therefore considered preferable in our analysis. The term \textit{action} is used because the resulting unit, $\frac{\text{joule-seconds}}{\text{token}}$, corresponds to the physical unit of action in physics. Additionally, the symbol $S$ was chosen as it is commonly used to denote action in physics.

\begin{figure}[!th]
    \centering
    \begin{subfigure}{0.43\textwidth}
        \includegraphics[width=\linewidth]{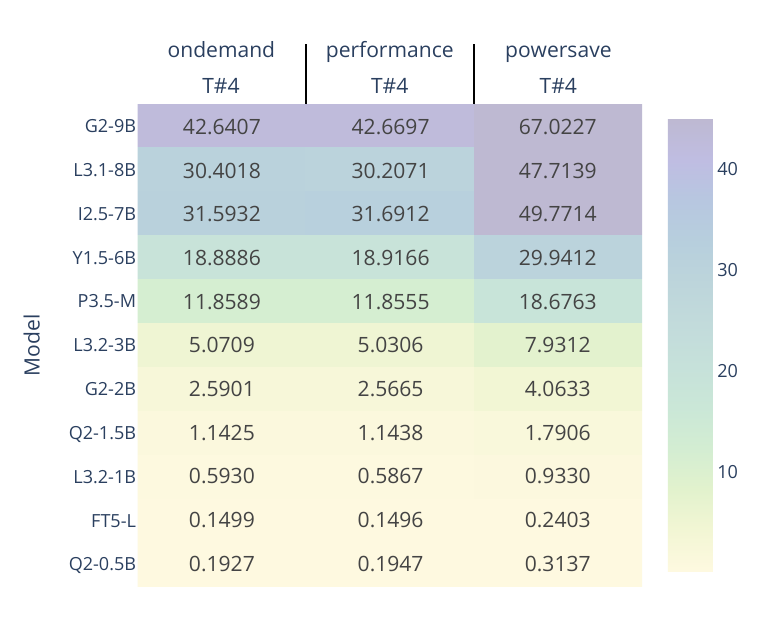}
        \caption{}
        \label{fig:prefill_action_qkm}
    \end{subfigure}
    \begin{subfigure}{0.43\textwidth}
        \includegraphics[width=\linewidth]{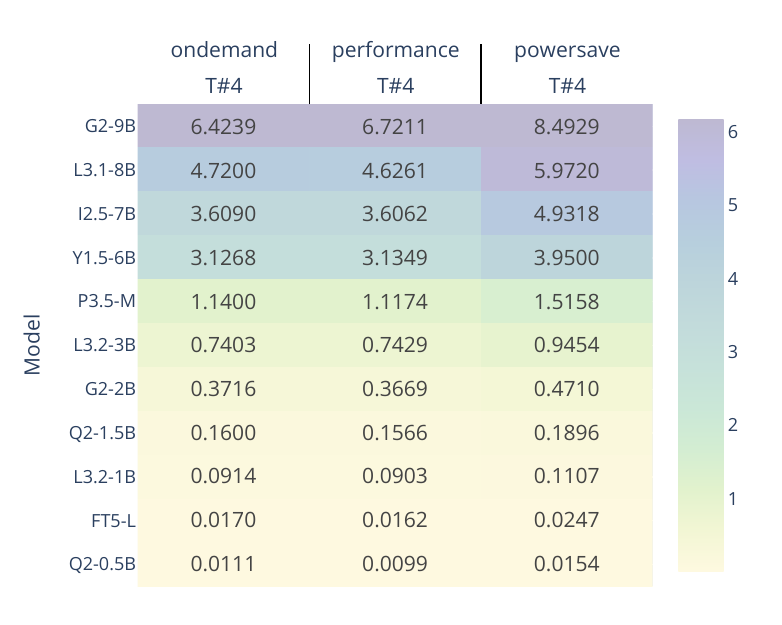}
        \caption{}
        \label{fig:prefill_action_q0}
    \end{subfigure}
    \caption{Comparison of generation action per token on RPi 5 for \qkm{} (a) and \qo{} (b).}
    \label{fig:prefill_action}
\end{figure}
\begin{figure}[!th]
    \centering
    \begin{subfigure}{0.9\textwidth}
        \centering
        \includegraphics[width=\textwidth]{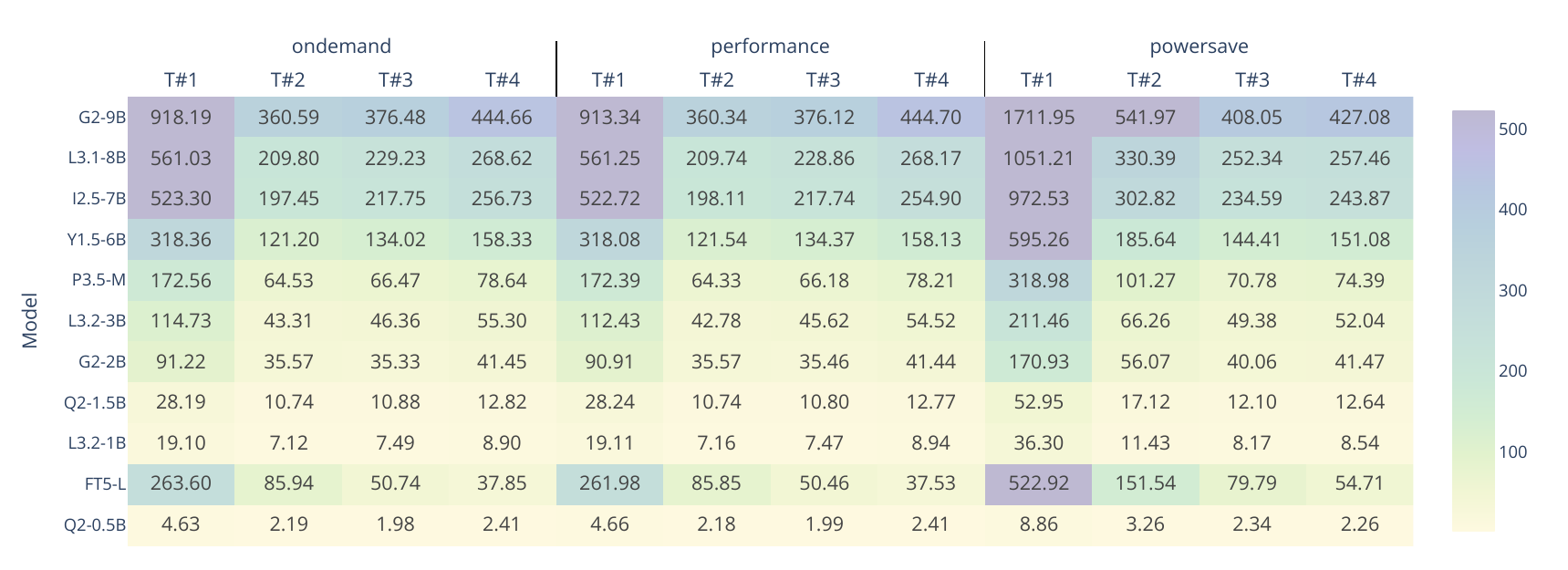}
        \caption{}
        \label{fig:gen_action_qkm}
    \end{subfigure}
    
    \begin{subfigure}{0.9\textwidth}
        \centering
        \includegraphics[width=\textwidth]{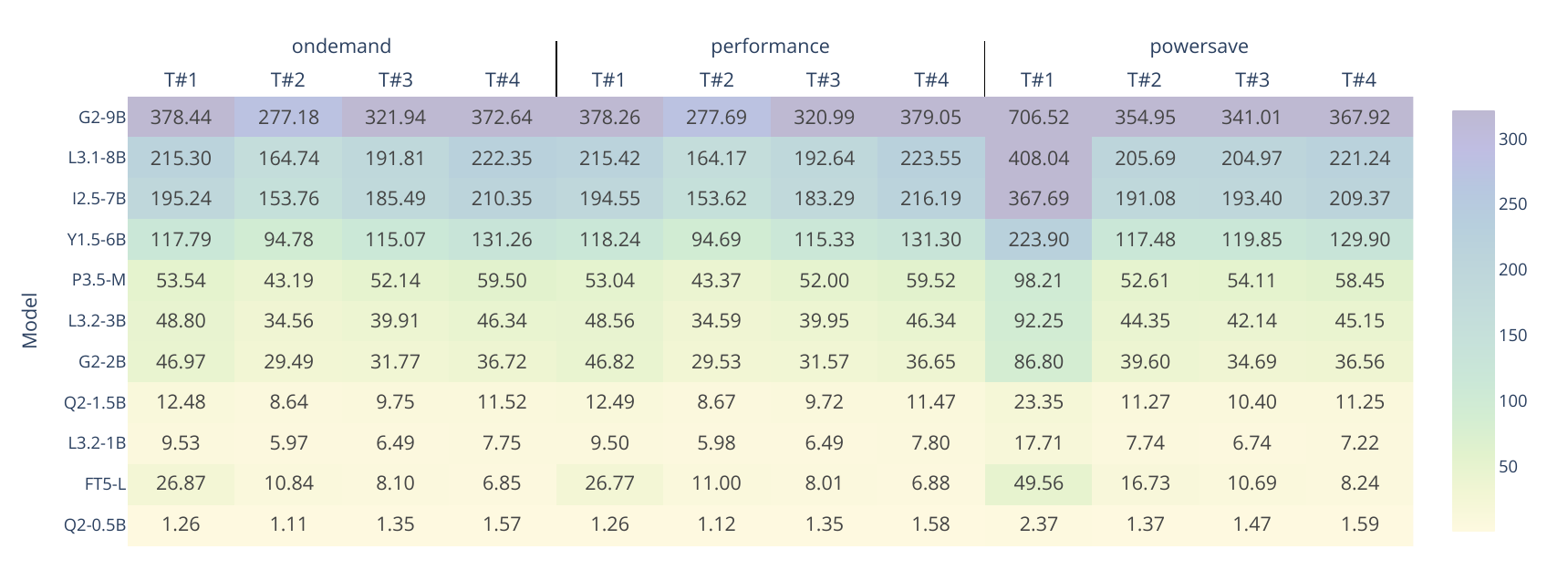}
        \caption{}
        \label{fig:gen_action_q0}
    \end{subfigure}
    
    \caption{Comparison of generation action per token on RPi 5 for \qkm{} (a) and \qo{} (b).}
    \label{fig:gen_action_comparison}
\end{figure}

When calculating $S_t$ for the prefill phase, Figure~\ref{fig:prefill_action} show that the \textit{ondemand} and \textit{performance} governors achieved a 36–38\% lower action per token than the \textit{powersave} governor for \qkm{} quantization. The same trend was observed for \qo{} quantization for which the \textit{ondemand} and \textit{performance} governors resulted in 15-31\% lower action per token. Only results for 4 threads are reported, as this configuration provided the best balance between energy consumption and token throughput across all models.

When analyzing the generation phase results in Figure~\ref{fig:gen_action_comparison}, it is evident that for \qkm{} quantization, the \textit{ondemand} and \textit{performance} governors also yielded the lowest action per token. For \qo{} quantization, for all models 2 threads were optimal with the only exception being Flan T5 Large for which 4 threads were optimal. \qkm{} quantization showed similar results with the only differences found for the models Gemma 2 2B and Qwen 2 0.5B, for which 3 threads minimized action per token. Overall, these results suggest that for decoder-only models, the  \textit{ondemand} and \textit{performance} governors perform with 2 threads perform best for most models, with some exceptions showing slightly better efficiancy with 3 threads.

\begin{figure}[!th]
    \centering
    
    \begin{subfigure}{0.247\textwidth}
        \centering
        \includegraphics[width=1\linewidth]{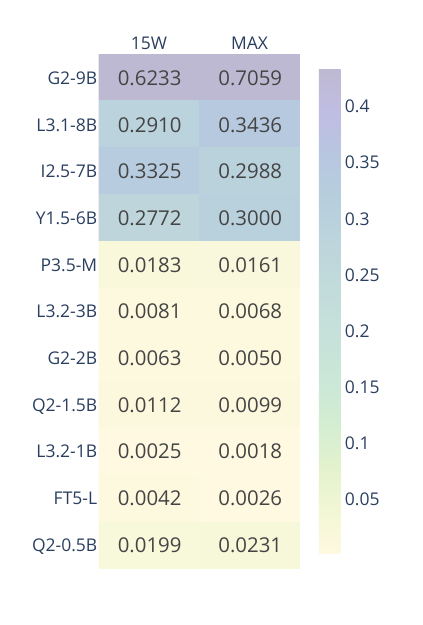}
        \caption{Prefill, \qkm{}}
        \label{fig:prefill_action_orin_km}
    \end{subfigure} \hfill
    \begin{subfigure}{0.247\textwidth}
        \centering
        \includegraphics[width=1\linewidth]{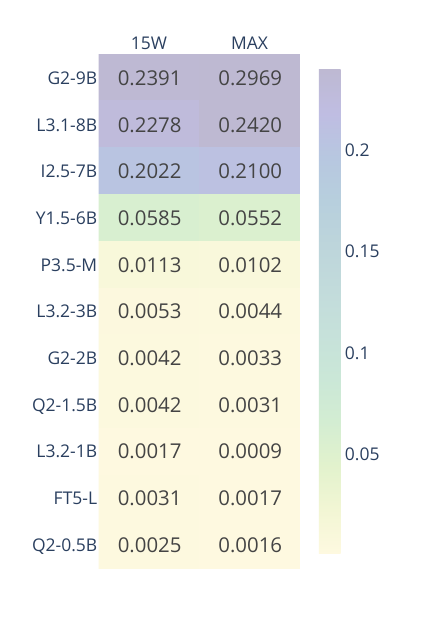}        
        \caption{Prefill, \qo{}}
        \label{fig:prefill_action_orin_0}
    \end{subfigure} \hfill
    \begin{subfigure}{0.228\textwidth}
        \centering
        \includegraphics[width=1\linewidth]{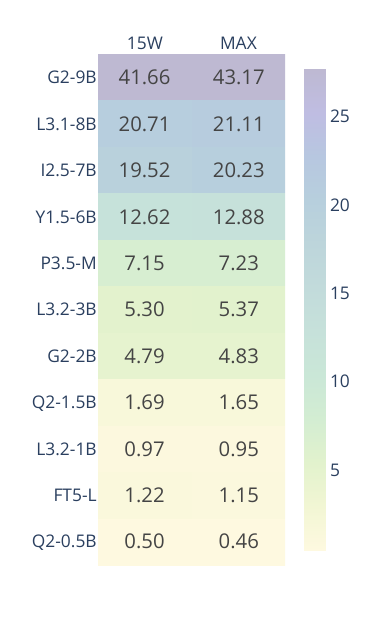}        
        \caption{Generation, \qkm{}}
        \label{fig:gen_action_orin_km}
    \end{subfigure} \hfill
    \begin{subfigure}{0.228\textwidth}
        \centering
        \includegraphics[width=1\linewidth]{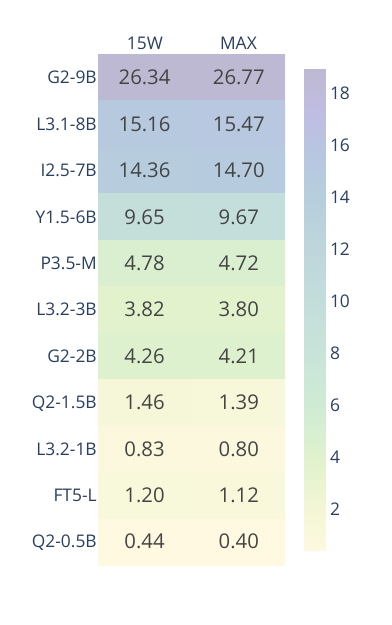}
        \caption{Generation, \qo{}}
        \label{fig:gen_action_orin_0}
    \end{subfigure} \hfill
  
    \caption{Action per token for prefill and generation phase for all power modes on Orin GPU}
    \label{fig:action_orin_overall}
\end{figure}

On the Orin GPU, the \textit{15W} mode resulted in the lowest energy consumption across all models, while the \textit{MAX} mode provided the highest inference speeds at the cost of increased energy usage. When using action per token as the decisive metric, it becomes evident that, for most smaller models, the \textit{MAX} mode offered the best energy-throughput trade-off in the prefill phase for both quantization schemes, while for the largest models \textit{15W} performed best. For \qo{} quantization, InternLM 2.5 7B was the first model, for which \textit{15W} performed better, contrarily to \qkm{} quantization wer this transition was already observable for Yi 1.5 6B. However, 2 outliers can be identified for \qkm quantization. For Qwen 2 0.5B despite being the smallest model, the \textit{15W} mode and for InterLM 2.5 7B despite being one of the larger models, the \textit{MAX} mode performed best.

In the generation phase, the \textit{MAX} mode yielded the optimal trade-off for all models up to Llama 3.2 3B with \qkm{} quantization and up to Phi 3.5 with \qo{} quantization. However, for larger models, the \textit{15W} mode proved to be more efficient. Results for the \textit{7W} mode were omitted, as it performed worse than the other two power modes in both token throughput and energy consumption.

Overall, a clear trend emerged: the \textit{MAX} mode provided better performance for smaller models, whereas the \textit{15W} mode was more efficient for larger models.

\begin{tcolorbox}[
    breakable,
    colback=blue!5!white, 
    colframe=blue!75!black, 
    title=Main Takeaways, 
    fonttitle=\small, 
    fontupper=\small,
    sharp corners, 
    boxrule=0.5pt, 
    left=2pt, right=2pt, top=2pt, bottom=2pt,
    before skip=6pt, after skip=2pt
]
\begin{itemize}
    \item For both prefill and generation phase on the RPi 5, the \textit{ondemand} and \textit{performance} CPU governors performed best with both quantization schemes. In the generation phase, 2 threads was the optimal thread count for most models.
    \item For GPU inference, the \textit{MAX} mode showed the optimal throughput-energy trade-off for most smaller models while for larger models the \textit{15W} mode showcased the best throughput-energy efficiency.
\end{itemize}
\end{tcolorbox}

\subsection{Usability Constraints and Practical Deployment Challenges}
\begin{figure}[!th]
    \centering
    
    \begin{subfigure}{0.48\textwidth}
        \centering
        \includegraphics[width=\linewidth]{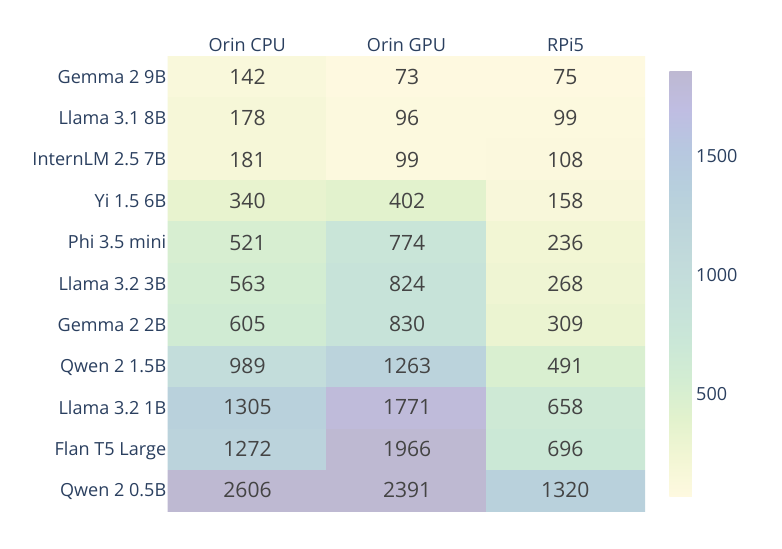}
        \caption{Runs}
        \label{fig:runs_battery}
    \end{subfigure}\hfill
    \begin{subfigure}{0.48\textwidth}
        \centering
        \includegraphics[width=\linewidth]{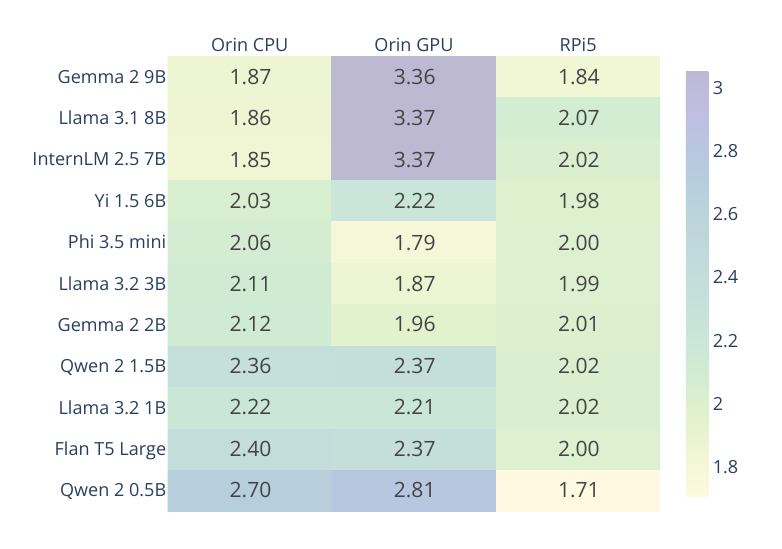}
        \caption{Time [hours]}
        \label{fig:hours_battery}
    \end{subfigure}\hfill
    
    \caption{Number of inference runs and Time [hours] until battery recharge for configurations with minimal energy consumption per run}
    \label{fig:battery_all}
\end{figure}

In the following, we analyze how inference speed and energy consumption impact the usability of LMs in real-world systems.

One key challenge is the long load times observed during the first run of each experiment and for all runs of larger models on the Orin. In any interactive use case, where users expect real-time responses, these high loading times—which translate to significant latencies—would likely degrade the user experience. To mitigate this, it is crucial to keep the model cached in memory between runs, as this substantially reduces load times, as discussed in Section~\ref{sec:results}. It can also be considered to pre-load models in idle times even before a request arrives to hide the latency from the user. 

Another potential limitation is the high memory demand of the models, even when they fit within the device's main memory. First, our results indicate that for the largest evaluated models, complications arose during GPU inference across both quantization schemes, as well as in CPU inference with \qo{} quantization, leading to significantly increased loading times and reduced prefill performance. Second, in real-world deployments, edge devices are often not dedicated solely to LM inference but may need to support additional applications. If the inference process occupies the majority of system memory, other applications may fail to run concurrently, leading to frequent memory paging. Since storage accesses are inherently time-consuming, this could cause substantial performance degradation.

For use cases where human users read generated tokens in real-time, reading speed serves as a baseline to determine whether token throughput is sufficient to maintain a smooth user experience. If tokens are generated slower than the average reading speed, users may need to pause frequently, leading to disruptions and a degraded experience \cite{liuAndesDefiningEnhancing2024}. The average reading speed of an English reader is approximately 4 words per second \cite{brysbaert_read_speed_2019}. According to OpenAI, one token corresponds to roughly 0.75 words \cite{tokens_to_words}. Based on this, an optimal generation throughput should be at least 5.3 $\frac{tokens}{s}$ to ensure a seamless experience. However, human reading speed varies significantly between individuals; therefore, this threshold should be considered only as an approximate guideline for assessing generation speed requirements in such use cases. When selecting the highest generation throughput for each model on the RPi 5, only models up to Llama 3.2 3B with \qkm{} quantization and up to Phi 3.5 mini with \qo{} quantization were able to exceed the 5.3 $\frac{tokens}{s}$ threshold. For Orin CPU inference, all models up to Yi 1.5 6B with \qkm{} quantization and up to InternLM 2.5 7B with \qo{} quantization met this requirement. However, for larger models, a degraded user experience seems likely due to insufficient generation throughput. In contrast, on the Orin with GPU inference, all models exceeded 5.3 $\frac{tokens}{s}$ for both quantization schemes, ensuring a smoother reading experience. 
 
However, generation speed requirements depend on the specific use case. In some scenarios, lower generation speeds may be sufficient, while in others, higher speeds could be advantageous. For instance, if an LM is summarizing documents as a background process—without user interaction or strict time constraints—slower generation speeds may be acceptable. In contrast, multi-agent LM systems could benefit from higher generation speeds, as LMs can typically process input tokens much faster than human reading speed.

As edge devices are often battery-powered or have limited energy availability, it is important to assess how many inference runs can be performed before the device runs out of battery. For each model, we used the configuration with the lowest mean energy consumption per run (considering runs 2–5 of an experiment) to estimate the number of inference runs that could be executed at 100\% utilization (i.e., with no idle time) on a fully charged smartphone battery. As a representative example, we selected the Samsung Galaxy S24 Ultra, a high-end smartphone with a nominal battery capacity of $18.84Wh=67824J$. As it is observable, even the maximum runtime remained below 3.5 hours, raising concerns about the feasibility of LM inference in energy-constrained environments, particularly when inference is performed frequently. That said, it is important to acknowledge the caveats of this analysis. First, our analysis considers only one input prompt, whereas different prompts could lead to variations in both the number of inference runs and total runtime before recharging. Second, the configurations minimizing total energy consumption were used for this evaluation. On the RPi 5, this often corresponds to the \textit{powersave} governor, which reduces inference speed. In real-world applications, this performance might be too slow, necessitating a less energy-efficient configuration. Because of this, these results should be interpreted as an upper bound on the number of feasible inference runs.

\begin{tcolorbox}[
    breakable,
    colback=blue!5!white, 
    colframe=blue!75!black, 
    title=Main Takeaways, 
    fonttitle=\small, 
    fontupper=\small,
    sharp corners, 
    boxrule=0.5pt, 
    left=2pt, right=2pt, top=2pt, bottom=2pt,
    before skip=6pt, after skip=2pt
]
\begin{itemize}
    \item It is doubtful whether generation speed of edge in inference is satisfying for the largest evaluated models for a wide range of use cases, especially with CPU inference
    \item However, performance for smaller models is likely satisfactory for many uses cases as generation throughput exceeded average human read speed even with CPU inference 
    \item Due to the high energy consumption, it is questionable whether highly frequent LM inference is feasible in energy-constrained environment
\end{itemize}
\end{tcolorbox}

\subsection{Cost Analysis}

With self-deployed models at the edge, no service costs are incurred for the user in contrast to the API costs of cloud services. For instance, at the time of writing, executing inference with the \textit{GPT-4o mini} model via the OpenAI API incurs a cost of 15\textcent{} per million input tokens and 60\textcent{} per million output tokens\footnote{"Ignoring Batch API prices and cached token prices"} \cite{openai_pricing}.

\begin{figure}[!th]
    \centering
    \includegraphics[width=0.8\linewidth]{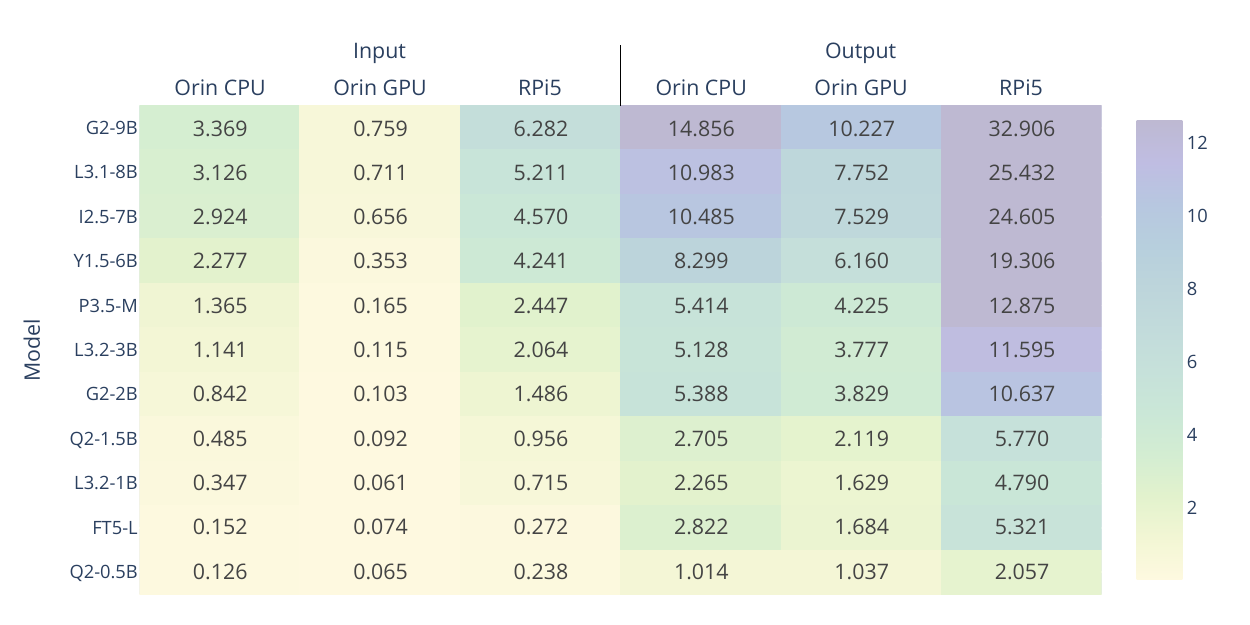}
    \caption{Energy cost per one million input and output tokens for different devices in cents}
    \label{fig:cost_per_million}
\end{figure}

However, edge deployments also incur operational costs, as energy consumption contributes to long-term expenses, in addition to the initial device purchase price. Using energy measurements from our experiments, we computed the cost per one million input and output tokens (illustrated in Figure~\ref{fig:cost_per_million}). The calculations assume an optimal thread count with the \textit{ondemand} governor for CPU inference, while for GPU inference, the \textit{15W} power mode was selected as the most energy-efficient setting. An electricity price of 20\textcent/kWh was chosen as a middle ground, considering that energy costs vary significantly by location and sector. For example, in the U.S., electricity prices in September 2024 ranged from under 10\textcent/kWh to over 40\textcent/kWh \cite{us_electricity}.

It can further be observed that the operational costs for one million input tokens range from 0.040\textcent{} for Llama 3.2 1B on GPU to 6.282\textcent{} for Gemma 2 9B on the RPi 5. Even in the most expensive scenario, this remains 2.39 times cheaper than the service cost of the OpenAI API, while in the best-case scenario, it is 375 times lower. For one million output tokens, the energy costs span from 1.014\textcent{} for Qwen 2 0.5B on the Orin CPU up to 32.906\textcent{} for Gemma 2 9B on the RPi 5. Even in the worst-case scenario, the cost remains 1.82 times lower than the OpenAI API service cost for GPT-4o mini, while in the best-case scenario, it is 59.17 times lower.

However, several caveats should be considered when interpreting these cost estimates. First, we do not account for power supply inefficiencies or energy usage during loading and idle states. Second, the operational costs presented here only consider energy consumption, whereas other expenses, such as device maintenance, may also contribute to the total cost. Third, these estimates do not factor in model quality, meaning that higher cloud service prices may be justified by more sophisticated model capabilities. Thus, to derive definitive conclusions, more precise energy measurements and a more comprehensive cost model incorporating additional operational factors are required.

Despite these considerations, the significantly lower operational costs—particularly for smaller models on GPU—suggest that self-deploying LMs at the edge could be more cost-efficient than cloud services in certain scenarios. This is especially relevant when idle edge resources can be leveraged, as this eliminates device acquisition costs. Moreover, self-deployment allows greater flexibility in model selection, enabling the deployment of the smallest model that meets task-specific requirements. Given that smaller models consume substantially less energy, as demonstrated in our results, this approach could yield further cost benefits. However, in scenarios where highly sophisticated model capabilities are essential, cloud services may still be the more economical option.

Beyond comparing edge inference with cloud services, it is also valuable to analyze the cost differences between the tested devices. As shown in Figure~\ref{fig:cost_per_million}, GPU inference on the Orin incurs significantly lower energy costs than CPU inference on the RPi 5. However, the initial device cost must also be considered. While the RPi 5 is much more affordable, the Orin's higher efficiency raises the question of when its purchase becomes justifiable from a cost perspective. Based on retail prices from DigiKey’s US store at the time of writing, the RPi 5 (including power supply and fan) costs \$97, whereas the Orin is priced at \$598.80—resulting in a price difference of \$501.80.

\begin{figure}[!t]
    \centering
    \includegraphics[width=\linewidth]{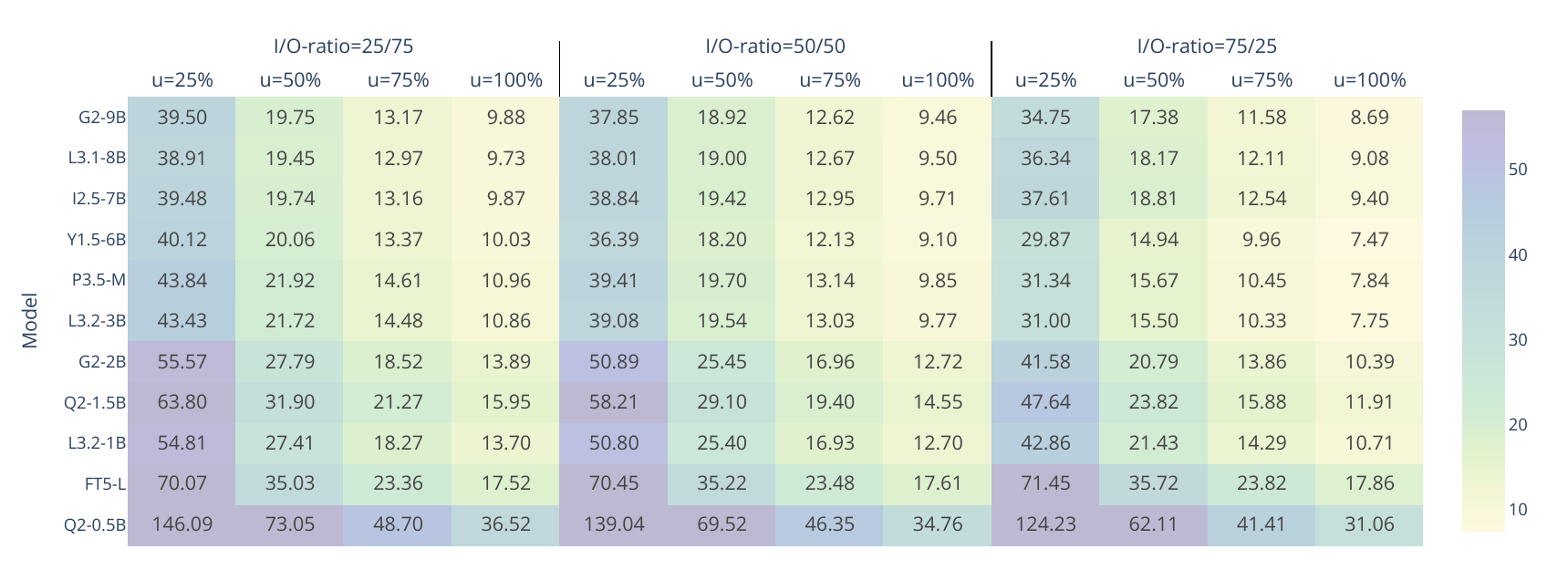}
    \caption{Break-even time in years between RPi 5 and Orin GPU for different IO-token ratios and device utilizations.}
    \label{fig:break_even}
\end{figure}

To better understand the cost-effectiveness of choosing the Orin over the RPi 5, we calculated break-even times for 12 scenarios with varying input/output token ratios and device utilization ($\mu$). The break-even time represents the duration needed for the operational cost savings of the Orin to offset its higher purchase price. For these calculations, we assumed an electricity price of 20\textcent{}/kWh. The RPi 5 was tested with the default governor (\textit{ondemand}) and throughput-optimal thread counts, while the Orin used the default power mode (\textit{15W}). Since \qo{} quantization proved more energy-efficient across both inference phases, we based our comparison on its results, excluding the energy consumption of the loading phase. As shown in Figure~\ref{fig:break_even}, input-heavy workloads result in shorter break-even times than output-heavy workloads, as the energy savings of GPU inference are more pronounced in the prefill phase than in the generation phase. Even with $\mu = 100\%$ utilization (i.e., continuous inference with no idle time), the fastest break-even scenario takes nearly 9 years. For balanced or output-heavy workloads, the required time increases even further, making it unlikely that the Orin would ever financially break even in typical edge AI deployments.

While these break-even time estimates provide insight, several caveats must be considered when interpreting the results. First, electricity prices have a direct inverse impact on break-even times. Doubling the electricity price would halve the time required to offset the Orin’s higher purchase cost. Second, computational performance is not accounted for in this analysis. In this respect, the higher throughput of the Orin may justify its higher purchase price for certain computationally demanding use cases. The utilization values in Figure~\ref{fig:break_even} are based on the Orin, which is the more powerful device. As a result, in some scenarios, the RPi 5 would require utilization exceeding 100\% to match the Orin’s performance, which is not feasible. Consequently, the break-even times for workloads corresponding to $\mu=100\%$ on the RPi 5 are significantly longer than the minimum values observed. Thus, for workloads where the RPi 5 provides sufficient performance, it remains the more cost-effective choice over the Orin.

\begin{tcolorbox}[
    breakable,
    colback=blue!5!white, 
    colframe=blue!75!black, 
    title=Main Takeaways, 
    fonttitle=\small, 
    fontupper=\small,
    sharp corners, 
    boxrule=0.5pt, 
    left=2pt, right=2pt, top=2pt, bottom=2pt,
    before skip=6pt, after skip=2pt
]
\begin{itemize}
    \item Edge inference could potentially yield cost benefits over cloud services due to its flexibility in model choice.
    \item From a pure cost perspective the RPi 5 is the more efficient device than the Orin.
\end{itemize}
\end{tcolorbox}

\section{Conclusions}
\label{sec:conclusion}
In this paper, we explored the feasibility and trade-offs of LM inference on edge devices, evaluating key performance factors such as memory usage, inference speed, and energy consumption across CPU- and GPU-based platforms. Our findings highlight the complex interplay between model size, hardware constraints, and quantization strategies, emphasizing the need for careful optimization to enable practical deployment in edge environments. While challenges remain with the current state-of-the-art technologies, our results also reveal promising opportunities, particularly in specific usability scenarios where local inference enhances responsiveness and autonomy, as well as potential cost benefits over cloud-based alternatives. We are optimistic that current trends, such as the emergence of GenAI-optimized Neural Processing Units (NPUs) \cite{HailoH10, QualcommNPU}, will further improve the state-of-the-art of LM inference at the edge, and that this study can contribute to advancing the R\&D landscape of LM inference-aware hardware optimizations, hardware-aware model optimizations, improved inference frameworks, and adaptive resource management, ultimately making edge-centric AI systems more efficient and scalable.

\bibliographystyle{ACM-Reference-Format}
\bibliography{bibliography}

\appendix
\end{document}